\documentclass{article}

\usepackage{microtype}
\usepackage{graphicx}
\usepackage{subfigure}
\usepackage[dvipsnames]{xcolor}
\usepackage{booktabs} %
\usepackage{multirow}

\usepackage{hyperref}

\usepackage[accepted]{icml2025}

\usepackage{amsfonts}
\usepackage{amssymb}
\usepackage{amsmath}
\usepackage{amsthm}
\usepackage{bm}
\usepackage{mathtools}
\mathtoolsset{showonlyrefs}

\usepackage{enumitem}
\usepackage{nicefrac}

\usepackage{mdframed}
\definecolor{theoremcolor}{rgb}{255, 255, 255}
\mdfsetup{
    backgroundcolor=theoremcolor,
    linewidth=1pt,
    leftline=true,
    topline=false,
    rightline=false,
    bottomline=false 
}

\newmdtheoremenv{definition}{Definition}
\newmdtheoremenv{proposition}{Proposition}
\newmdtheoremenv{corollary}{Corollary}
\newmdtheoremenv{theorem}{Theorem}
\newmdtheoremenv{lemma}{Lemma}
\newmdtheoremenv{example}{Example}

\def\RR{\mathbb{R}}
\def\EE{\mathbb{E}}

\def\cB{\mathcal{B}}
\def\cC{\mathcal{C}}

\def\cF{\mathcal{F}}
\def\cL{\mathcal{L}}

\def\cP{\mathcal{P}}
\def\cS{\mathcal{S}}
\def\cV{\mathcal{V}}
\def\cX{\mathcal{X}}
\def\cY{\mathcal{Y}}
\def\cW{\mathcal{W}}

\def\ones{\bm{1}}

\def\u{{\bm{u}}}
\def\v{{\bm{v}}}
\def\w{{\bm{w}}}
\def\x{{\bm{x}}}
\def\y{{\bm{y}}}
\def\muv{{\bm{\mu}}}
\def\thetav{{\bm{\theta}}}
\def\tauv{{\bm{\tau}}}

\def\U{{\bm{U}}}

\DeclareMathOperator*{\argmax}{argmax}

\DeclareMathOperator*{\dom}{dom}
\DeclareMathOperator*{\conv}{conv}

\def\LSE{\mathrm{LSE}}
\def\MLE{\mathrm{MLE}}
\def\sigmoid{\mathrm{sigmoid}}
\def\softplus{\mathrm{softplus}}
\def\softmax{\mathrm{softmax}}

\def\aka{a.k.a.\ }
\def\wrt{w.r.t.\ }

\def\mytitle{Joint Learning of Energy-based Models and their Partition Function}
\icmltitlerunning{\mytitle}

\begin{document}

\twocolumn[
\icmltitle{\mytitle}

\icmlsetsymbol{equal}{*}

\begin{icmlauthorlist}

\icmlauthor{Michaël E. Sander}{gdm}
\icmlauthor{Vincent Roulet}{gdm}
\icmlauthor{Tianlin Liu}{gdm}
\icmlauthor{Mathieu  Blondel}{gdm}
\end{icmlauthorlist}

\icmlaffiliation{gdm}{Google DeepMind}

\icmlcorrespondingauthor{Mathieu Blondel}{mblondel@google.com}
\icmlcorrespondingauthor{Michaël E. Sander}{michaelsander@google.com}

\icmlkeywords{ebms, duality, mle}

\vskip 0.3in
]

\printAffiliationsAndNotice{}  %

\begin{abstract}
Probabilistic energy-based models (EBMs) offer a flexible framework for parameterizing probability distributions using neural networks.
However, learning probabilistic EBMs by exact maximum likelihood estimation (MLE) is generally intractable, due to the need to compute the partition function (normalization constant).
In this paper, we propose a novel formulation for approximately learning probabilistic EBMs in combinatorially-large discrete spaces, such as sets or permutations. 
Our key idea is to jointly learn both an energy model and its log-partition, both parameterized as neural networks.
Our approach not only provides a novel tractable objective criterion to learn EBMs by stochastic gradient descent (without relying on MCMC), but also a novel means to estimate the log-partition function on unseen data points.
On the theoretical side, we show that our approach recovers the optimal MLE solution when optimizing in the space of continuous functions.
Furthermore, we show that our approach naturally extends to the broader family of Fenchel-Young losses, allowing us to obtain
the first tractable method for optimizing the sparsemax loss in combinatorially-large spaces.
We demonstrate our approach on multilabel classification and label ranking.
\end{abstract}

\section{Introduction}

Probabilistic energy-based models (EBMs) are a powerful framework for parameterizing probability distributions using neural networks, without any factorization assumptions. They have been successfully used both in the unsupervised setting (density estimation, image generation), where the goal is to learn a probability density function $p(\x)$ and in the supervised learning setting (classification, structured prediction), where the goal is to learn a conditional probability distribution $p(\y|\x)$.
Probabilistic EBMs include Bolzmann machines \citep{ackley1985learning} as a special case.

Unfortunately, this flexibility makes learning, sampling and inference with probabilistic EBMs much more challenging. Indeed, without factorization assumptions, turning an energy function into a valid probability distribution involves computing the partition function (normalization constant), which is generally intractable to compute exactly in continuous spaces or in combinatorially-large discrete spaces.
To perform maximum likelihood estimation (MLE) of the network parameters,
a standard approach is to rely on Markov chain Monte Carlo (MCMC) methods to estimate stochastic gradients of the log-likelihood function \citep{song2021train}. 
However, deriving an MCMC sampler is usually case by case. For instance, for continuous outputs we may use Langevin-MCMC and for binary outputs we may use Gibbs sampling.
In practice, as is the case in contrastive divergences \citep{hinton2002training}, MCMC iterations are not run to convergence, leading to biased stochastic gradient estimates \citep{fischer2010empirical}. In continuous spaces, score matching \citep{hyvarinen2005estimation} is often used as an alternative to MLE but it requires computing the Laplacian operator, which is challenging in high dimension.
In combinatorially-large discrete spaces, one can use the generalized perceptron loss \citep{lecun2006tutorial} or the generalized Fenchel-Young loss \citep{gfy}, which only involve computing an argmax or a regularized argmax, respectively. These losses sidestep computing the partition function, but they learn non-probabilistic EBMs.

In this paper, we propose a novel formulation for learning probabilistic EBMs $p(\y|\x)$ in combinatorially-large discrete spaces (such as sets or permutations in our experiments) by MLE or any loss function in the Fenchel-Young family \citep{blondel2020learning}. Our approach not only provides a new tractable objective criterion that can be optimized by stochastic gradient descent (SGD) without MCMC, but also a new means to estimate the log-partition function on unseen data points using a jointly-learned separate neural network.
We demonstrate our method on multilabel classification (space of sets) and label ranking (space of permutations).

\paragraph{Contributions.} 

\begin{itemize}[topsep=0pt,itemsep=3pt,parsep=3pt,leftmargin=16pt]

\item After reviewing some extensive background, we propose a novel min-min formulation for learning probabilistic EBMs $p(\y|\x)$ in combinatorially-large discrete spaces. Our approach consists in jointly learning two separate functions: an energy model and its log-partition.

\item When minimizing the proposed objective in the space of continuous functions, we show that our approach exactly recovers MLE. 

\item In practice, we propose to parameterize the log-partition as a neural network and we replace minimization in the space of functions with minimization in the space of network parameters. An advantage of our approach is that the learned log-partition network can benefit from the universal approximation properties of neural networks, as we empirically demonstrate.

\item To jointly learn the energy model and the log-partition, we propose an MCMC-free doubly stochastic optimization scheme. 
Our stochastic gradients are unbiased: our only assumption is that we are able to sample from a reference distribution, which can be the uniform distribution when no prior knowledge is available.

\item We generalize our approach to the broader family of Fenchel-Young losses \citep{blondel2020learning}, such as the sparsemax loss \citep{martins2016softmax}. In this case, our approach consists in jointly learning the energy-based model and a Lagrange mutiplier (dual variable), that we parametrize as a neural network in practice. This is to our knowledge the first tractable approach for learning EBMs with the sparsemax loss in general combinatorially-large discrete spaces.
\end{itemize}
\vspace{-1.em}
\paragraph{Notation.}
We denote the continuous input space by $\cX$ and the discrete output set by $\cY$.
We denote the set of continuous functions $g \colon \cX \times \cY \to \RR$ as $\cF(\cX \times \cY)$. Similarly, we use $\cF(\cX)$ for continuous functions $\tau \colon \cX \to \RR$. We denote the set of conditional positive measures over $\cY$ conditioned on $\cX$ as $\cP_+(\cY|\cX)$. Similarly, we use $\cP_1(\cY|\cX)$ for conditional probability measures. Given $\x \in \cX$, we denote the partial evaluation of $g \in \cF(\cX \times \cY)$ as $g(\x, \cdot) \in \cF(\cY)$.
Given $h \in \cF(\cY)$ and $p \in \cP_+(\cY)$, we define
$\langle h, p \rangle \coloneqq \sum_{\y \in \cY} h(\y) p(\y)$. We denote the convex conjugate of $f(u)$ by $f^*(v) \coloneqq \sup_u uv - f(u)$. Throughout this paper, we will use $\EE_{(\x,\y)}$ as a shorthand for $\EE_{(\x,\y) \sim \rho_{\cX\times\cY}}$ where $\rho_{\cX\times\cY}$ is the joint data distribution over $\cX \times \cY$ and 
$\EE_\x$ as a shorthand for 
$\EE_{\x \sim \rho_{\cX}}$,
where $\rho_\cX$ is the marginal data distribution, i.e.,
$\rho_\cX(\x) \coloneqq \sum_{\y \in \cY} \rho_{\cX\times\cY}(\x, \y)$.

\section{Background}

\subsection{Probabilistic energy-based models (EBMs)}
\label{sec:proba_ebm}

Given a function $g \in \cF(\cX \times \cY)$, 
which captures the affinity between an input $\x \in \cX$ and an output $\y \in \cY$ (typically, a parameterized neural network),
we define a probabilistic energy-based model (EBM) as
\begin{equation}
\label{eq:ebm}
p_g(\y|\x) \coloneqq \frac{q(\y|\x)\exp(g(\x, \y))}{\sum_{\y' \in \cY}q(\y'|\x)\exp(g(\x, \y'))},
\end{equation}
where $q \in \cP_+(\cY|\cX)$ is a prior conditional positive measure.
When such a measure is not available, we can just use the uniform positive measure, which gives
\begin{equation}
p_g(\y|\x) = \frac{\exp(g(\x, \y))}{\sum_{\y' \in \cY}\exp(g(\x, \y'))}.
\end{equation}
We focus on \eqref{eq:ebm} for generality, though in our experiments we will use a uniform reference measure. Throughout this paper, we leave the dependence on $q$ implicit, as it is a fixed reference measure.
It is well-known \citep{boyd2004convex} that \eqref{eq:ebm} can be written from a variational perspective as
\begin{equation}
\label{eq:softargmax_variational}
p_g(\cdot|\x) = \argmax_{p \in \cP_1(\cY)} \langle g(\x, \cdot), p \rangle - \mathrm{KL}(p, q(\cdot|\x)).
\end{equation}
The corresponding log-partition, \aka log-sum-exp, is
\begin{equation}
\label{eq:lse}
\LSE_g(\x) \coloneqq \log \sum_{\y' \in \cY} q(\y'|\x) \exp(g(\x,\y')).
\end{equation}
It is related to the log-likelihood of the pair $(\x,\y)$ through
\begin{equation}
\label{eq:log_likelihood}
\log p_g(\y|\x) = g(\x, \y) - \LSE_g(\x) + \log q(\y|\x).
\end{equation}
Similarly to \eqref{eq:softargmax_variational}, we have
\begin{equation}
\label{eq:lse_variational}
\LSE_g(\x) = \max_{p \in \cP_1(\cY)} \langle g(\x, \cdot), p \rangle - \mathrm{KL}(p, q(\cdot|\x)).
\end{equation}

\paragraph{Architecture of EBMs.}

As $g$ is an affinity function, it is both natural and standard to decompose it as
\begin{equation}
\label{eq:g_decomposition}
g(\x, \y) \coloneqq \Phi(h(\x), \y),
\end{equation}
where 
$\Phi(\thetav, \y)$ is a coupling function and
$\thetav \coloneqq h(\x)$ is a model function producing logits
$\thetav \in \Theta$ \citep{gfy}. 
Typically, $h$ and $\Phi$ are designed on a per-task basis. 
The simplest example of coupling is the bilinear form
\begin{equation}
\label{eq:bilinear_coupling}
\Phi(\thetav, \y) \coloneqq \langle \thetav, \y \rangle.
\end{equation}
In this case, an EBM coincides with an exponential family distribution with natural parameters $\thetav$.
We will define more advanced couplings in Section \ref{sec:experiments}.

\subsection{Inference with EBMs}

\paragraph{Computing the mode.}

Predicting the most likely output associated with an input $\x$ according to an EBM corresponds to  computing the mode of the learned probability distribution, defined as
\begin{equation}
\label{eq:argmax}
\y^\star_g(\x) 
\coloneqq \argmax_{\y \in \cY} p_g(\y|\x) 
= \argmax_{\y \in \cY} q(\y|\x)\exp(g(\x, \y)).
\end{equation}
Unfortunately, when $\cY$ is a combinatorially-large discrete set, this problem is often intractable. A common approximation is to replace $\cY$ with a convex set $\cC \supset \cY$. 
The tightest such set is $\cC = \conv(\cY)$, the convex hull of $\cY$.
For example, in multilabel classification with $k$ labels, the output space is the power set of $[k]$, which can be represented as $\cY = \{0,1\}^k$ and whose convex hull is $\conv(\cY) = [0,1]^k$.

With uniform prior $q$, 
assuming that $g \in \cF(\cX \times \cC)$, 
and not just $g \in \cF(\cX \times \cY)$,
we can solve the relaxed problem
\begin{equation}
\label{eq:relaxed_argmax}
\x \mapsto \argmax_{\muv \in \cC} g(\x, \muv) \approx \y^\star_g(\x),
\end{equation}
where we used that $\exp$ is monotonically increasing.
If the coupling $\Phi(\thetav, \muv)$ is concave in $\muv$, this problem can be solved optimally in polynomial time.
Typically, one can use a gradient-based solver to obtain an approximate solution in $\cC$ and then use a rounding procedure to produce a prediction that belongs to $\cY$.
If $\Phi(\thetav, \muv)$ is the bilinear coupling \eqref{eq:bilinear_coupling},
then \eqref{eq:relaxed_argmax} is known as maximum a posteriori (MAP) inference \citep{wainwright2008graphical} when $\cC = \conv(\cY)$. According to the fundamental theorem of linear programming, a solution happens at one of the vertices of $\cC$ and therefore $\y^\star_g(\x) \in \cY$ in this case.

\paragraph{Computing the mean.}

Another important problem is that of computing the conditional expectation
\begin{equation}
\label{eq:conditional_expectation}
\muv_g(\x) 
\coloneqq \EE_{\y \sim p_g(\cdot|\x)}[\y]
= \sum_{\y \in \cY} p_g(\y|\x) \y
\in \conv(\cY).
\end{equation}
In the special case of the bilinear coupling \eqref{eq:bilinear_coupling},
we have that $\muv_g(\x)$ is the gradient \wrt $h$ of $\LSE_g(\x)$.
Moreover, \eqref{eq:conditional_expectation} is known as marginal inference and
$\conv(\cY)$ is known as the marginal polytope, because,
if $\cY$ uses a binary encoding (indicator function) of outputs,
then $\muv_g(\x)$ contains marginal probabilities \citep{wainwright2008graphical}. For example, if $\cY = \{0,1\}^k$, then $[\muv_g(\x)]_j$ contains the marginal probability of label $j$ given $\x$. 

When using the bilinear coupling \eqref{eq:bilinear_coupling} and
when the variables in $\cY$ form a tree, 
$\LSE_g$ and $\muv_g$ can be computed by message-passing-like dynamic programming algorithms \citep{wainwright2008graphical,edpbook}.
Efficient algorithms also exist for particular structures, such as spanning trees \citep{zmigrod2021efficient}.
However, these algorithms no longer work for non-bilinear couplings.
In addition, even in the case of bilinear couplings,
computing $\LSE_g$ and $\muv_g$ is intractable in general. For example, if $\cY$ is the space of permutations, then computing $\LSE_g$ and $\muv_g$ is known to be \#P-complete, making exact MLE intractable.  

\subsection{MLE of EBMs is intractable in general}

The log-likelihood of an observed pair $(\x,\y)$ is given in \eqref{eq:log_likelihood}.
Accordingly, the MLE objective,
which corresponds to expected risk minimization with a logistic loss,
is
\begin{align}
\cL_{\MLE}(g) 
&\coloneqq \EE_{(\x,\y)} \left[-\log p_g(\y|\x)\right]\\
&= \EE_{\x} \left[\LSE_g(\x)\right]
- \EE_{(\x,\y)} \left[g(\x, \y) -\log(q(\y|\x))\right] \label{eq:mle}.
\end{align}
Unfortunately, when $\cY$ is a combinatorially-large discrete set or a continuous set, this objective is intractable due to the log-partition function over all possible outputs.

\subsection{Approximate MLE of EBMs with MCMC}

Let us denote by $g_\w$ a function with parameters $\w$. Then,
the gradient can be computed by \citep{song2021train}
\begin{align}
\nabla_\w \cL_{\MLE}(g_\w) 
= 
&\EE_\x\left[
\EE_{\y' \sim p_{g_\w}(\cdot|\x)}[\nabla_\w g_\w(\x, \y')]
\right] - \\
&\EE_{(\x,\y)} \left[\nabla_\w g_\w(\x, \y)\right].
\end{align}
For completeness, see Appendix \ref{proof:gradient_estimator} for a proof.
Therefore, as long as we can draw samples from 
$p_{g_\w}(\cdot|\x)$,
we can have access to an unbiased Monte-Carlo estimate of the gradient, allowing us to perform optimization based on stochastic gradient descent. However, sampling from
$p_{g_\w}(\cdot|\x)$
is difficult.
The literature has focused on Markov chain Monte Carlo (MCMC) methods, such as Metropolis-Hastings, Gibbs sampling and Langevin. Typically, running MCMC until convergence is expensive. Therefore, MCMC is usually run for a small number of iterations, as in contrastive divergences \citep{hinton2002training}.
Unfortunately, truncated MCMC can lead to biased gradient updates that hurt the learning dynamics \citep{song2021train}. There are methods for bias removal but they typically greatly increase the variance.

\subsection{Non-MLE probabilistic approaches}

Departing from MLE, we can learn probabilistic EBMs of the form \eqref{eq:ebm}
using score matching \citep{hyvarinen2005estimation}.  It trains a model to
estimate the gradient of the log-density function of the data, known as the
score function.  However, score matching requires the log-density of the data
distribution to be continuously differentiable and finite everywhere, which
limits it to continuous output spaces. In addition, score matching requires to
compute and differentiate the Laplacian operator (the trace of the Hessian),
which is often challenging in high dimension. Even though some extensions
to discrete spaces have been proposed, such as concrete score matching
\citep{meng2022concrete} and ratio matching \citep{hyvarinen2007some},
score matching is seldom used in discrete spaces.

Another non-MLE approach is noise contrastive estimation (NCE) \citep{Pihlaja2010Family, gutmann2010noise,gutmann2012bregman}. It works by training a model to discriminate between true data samples and generated noise samples. However, the performance of NCE can be sensitive to the choice of the noise distribution.
In addition, the ratio of noise samples to data samples can affect performance and finding the optimal ratio usually requires experimentation.

\subsection{Loss functions for non-probabilistic EBMs}

When the goal is not to learn the probabilistic model \eqref{eq:ebm} but just the relaxed argmax \eqref{eq:relaxed_argmax}, we can use the generalized perceptron loss \citep{lecun2006tutorial},
\begin{equation}
(\x, \y) \mapsto 
\max_{\muv \in \cC} g(\x, \muv) - g(\x, \y).
\end{equation}
To ensure uniqueness of the argmax, we can add regularization $\Omega$ in \eqref{eq:relaxed_argmax} to define the regularized relaxed problem
\begin{equation}
\x \mapsto \argmax_{\muv \in \cC} g(\x, \muv) - \Omega(\muv).
\end{equation}
The corresponding generalized Fenchel-Young loss \citep{gfy} is
\begin{equation}
(\x, \y) \mapsto 
\max_{\muv \in \cC} g(\x, \muv) - \Omega(\muv) - g(\x, \y) + \Omega(\y).   
\end{equation}
Crucially, the gradient \wrt the parameters of $g$ can be computed easily using the envelope theorem. However, these losses do not learn probabilistic models. They only learn to make the (relaxed) argmax predict the correct output. See also the discussion on distribution-space vs.\ mean-space losses in \citet{blondel2020learning}.

\subsection{Log-partition as an optimization variable}

The idea of treating the log-partition as an optimization variable, rather than
as a quantity to compute or estimate, was also explored in several earlier works
in the unsupervised setting
\citep{Wang2018LearningTransdimensional,Arbel2021Generalized,
Senetaire2025LearningEBM}.  Compared to these works, our paper provides new
theoretical guarantees regarding MLE recovery when optimizing in function space
(Propositions \ref{prop:min_min_mle} and \ref{prop:finite}), extends to the
broader Fenchel-Young losses (Proposition \ref{prop:min_min_fy}), focuses on the
supervised structured prediction setting, and
most importantly, parameterizes the input-dependent log-partition as a neural
network with generalization ability (in the unsupervised setting, the
log-partition is an input-independent scalar value, therefore parameterizing the
log-partition as a neural network would not make sense).

\section{Proposed approach}

In this section, we first propose a novel min-min formulation for learning probabilistic EBMs by MLE. We then generalize it to the family of Fenchel-Young losses.

\subsection{A min-min formulation for MLE in EBMs}

\paragraph{Overview of the approach.}

The partition function (normalization constant) in \eqref{eq:ebm} can equivalently be formulated as an (intractable) equality constraint $\sum_{\y \in \cY} p_g(\y|\x) = 1$ for all $\x \in \cX$.
To deal with that constraint, our key idea is to introduce a Lagrange multiplier (dual variable), that we treat as a separate function $\tau \in \cF(\cX)$ to minimize over. The Lagrange multiplier exactly coincides with the log-partition function in the MLE (logistic loss) case.
Our approach consists in jointly learning the energy function $g \in \cF(\cX \times \cY)$
and the log-partition function $\tau \in \cF(\cX)$.
Using duality arguments in the space of functions (see Proposition \ref{prop:min_min_mle} below), we arrive at the objective function
\begin{equation}
\label{eq:proposed_mle}
\cL_{\MLE}(g, \tau) 
\coloneqq \EE_\x\left[L_{g,\tau}(\x)\right]
- \EE_{(\x,\y)} \left[g(\x, \y)\right],
\end{equation}
where
\begin{equation}
\label{eq:L_g_tau}
L_{g,\tau}(\x)
\coloneqq
\tau(\x) + \sum_{\y' \in \cY} q(\y'|\x)\left( \exp(g(\x, \y') - \tau(\x)) - 1\right).
\end{equation}
The minimization problem in the space of functions is then
\begin{equation}
\min_{g \in \cF(\cX\times\cY)} \min_{\tau \in \cF(\cX)}
\cL_{\MLE}(g, \tau).
\end{equation}
Intuitively, by minimizing the proposed objective with respect to both $g$ and
$\tau$, we are able to approximate the true
log-partition function while simultaneously learning the energy function so as
to fit the data.
In practice, we parameterize the energy model as $g_\w$ and the log-partition as $\tau_\v$. The minimization problem in the space of parameters is then
\begin{equation}
\min_{\w \in \cW} \min_{\v \in \cV}
\cL_{\MLE}(g_\w, \tau_\v).  
\end{equation}
In our experiments in Section \ref{sec:experiments}, we consider linear models, multilayer pecerptrons (MLPs) and residual networks (ResNets) for $g_\w$, and we consider constant models, MLPs, ResNets and input-convex neural networks (ICNNs) \citep{amos2017input} for $\tau_\v$.

\paragraph{Doubly stochastic gradient descent.}

When $q$ is a probability distribution (if such a distribution is not available, we simply use the uniform distribution), we can rewrite \eqref{eq:L_g_tau} as
\begin{equation}
\label{eq:L_g_tau_expectation}
L_{g,\tau}(\x)
=
\tau(\x) + \EE_{\y' \sim q(\cdot|\x)} \left[ \exp(g(\x, \y') - \tau(\x))\right] - 1.
\end{equation}
As a result, we can estimate $\cL_{\MLE}(g, \tau)$ and its stochastic gradients, 
provided that we can sample from $q(\cdot|\x)$ for any $\x$.
This suggests a doubly stochastic scheme, in which we sample both $(\x,\y)$ pairs from the data distribution $\rho_{\cX\times\cY}$ and $\y'$ prior samples from the reference distribution $q(\cdot|\x)$; see Algorithm \ref{algo:doubly_stochastic}. For simplicity, we group $\w \in \cW$ and $\v \in \cV$ as a tuple and optimize both blocks of parameters simultaneously using Adam \citep{kingma2014adam}. Although not explored in this work, it would also be possible to perform alternating minimization \wrt $\w \in \cW$ and $\v \in \cV$.

\begin{algorithm}[ht]
\centering
    \caption{Doubly stochastic objective value computation}
\begin{algorithmic}
\STATE \textbf{Inputs:} models $g$ and $\tau$, batch sizes $B$ and $B'$
\STATE Draw $(\x_1, \y_1), \dots, (\x_B, \y_B) \overset{\text{i.i.d.}}{\sim} \rho_{\cX \times \cY}$
\STATE Draw $\y'_{i, 1}, \dots, \y'_{i, B'} \overset{\text{i.i.d.}}{\sim} q(\cdot|\x_i)$ for $i \in [B]$
\STATE \small $\tilde{L}_{g,\tau}(\x_i)
\coloneqq \tau(\x_i) + \frac{1}{B'} \sum_{j=1}^{B'} (\exp(g(\x_i, \y'_{i,j}) - \tau(\x_i)) - 1)$
\STATE \textbf{Output:} 
$\frac{1}{B} \left(\sum_{i=1}^{B} \tilde{L}_{g,\tau}(\x_i) - g(\x_i, \y_i)\right)$
\end{algorithmic}
\label{algo:doubly_stochastic}
\end{algorithm}

\paragraph{Equivalence with MLE.}

\begin{figure*}[t]
    \centering
    \subfigure[Loss (train)]{\includegraphics[width=0.30\textwidth]{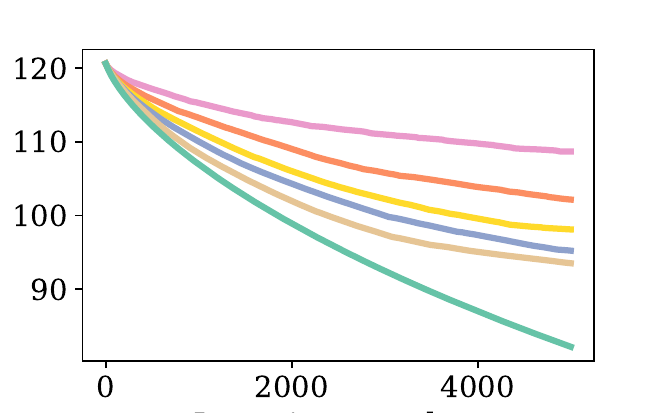}}
    \subfigure[Gradient norms (train)]{\includegraphics[width=0.30\textwidth]{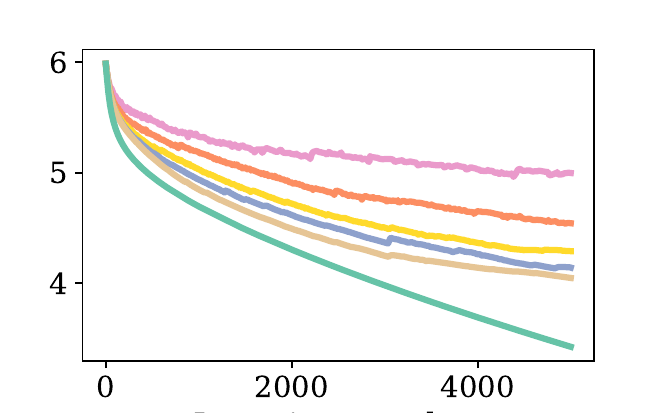}} 
    \subfigure[$f_1$-score (test)]{\includegraphics[width=0.30\textwidth]{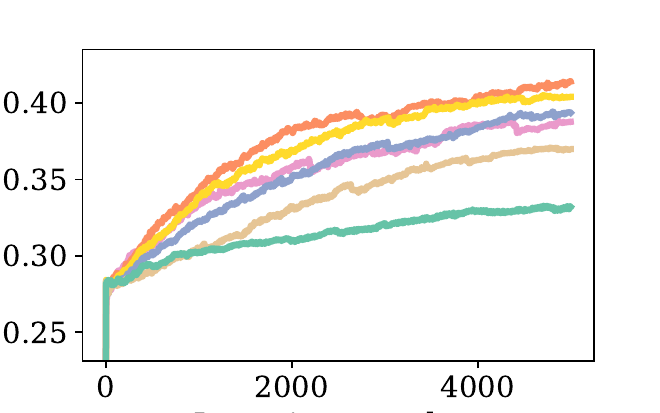}}
    \\
    \vspace{-.5em}
    \subfigure{\includegraphics[width=0.45\textwidth]{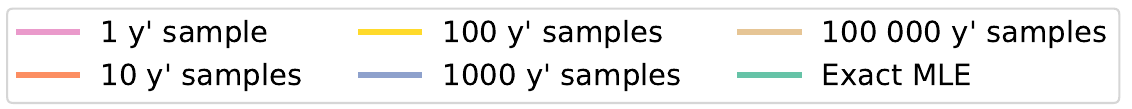}}
     \vspace{-.8em}
     \caption{{\bf Convergence of the proposed approach} as a function of the number of iterations, when varying the number of prior samples $\y'$ drawn.
     To be able to compute the exact MLE objective \eqref{eq:mle_finite_sum}, we use the unary multilabel model (Section \ref{sec:multilabel}) on the cal500 dataset (174 classes and therefore $2^{174}$ possible configurations) as the test bed. Note that the loss and gradient in plots (a) and (b) are computed using \eqref{eq:mle_finite_sum} even for our method. We make two key observations: i) our approach converges to exact MLE as predicted by our theory, ii) the number of $\y'$ samples can have a regularization effect on the test $f_1$-score. See Figure \ref{fig:learning_curves_all} for more results.}\label{fig:learning_curves_cal500}
\end{figure*}

Although our proposed objective \eqref{eq:proposed_mle} appears quite different from the original MLE (expected risk) objective,
we can show that they are in fact equivalent when minimizing over the space of functions.
\begin{proposition}{Optimality of min-min, MLE case}
\label{prop:min_min_mle}

Suppose that for all $\y \in \cY$, $\x \mapsto q(\y|\x)$ is continuous. Then, we have
\begin{equation}
\min_{g \in \cF(\cX\times\cY)} \cL_{\MLE}(g)
=\min_{\substack{g \in \cF(\cX\times\cY)\\\tau \in \cF(\cX)}} \cL_{\MLE}(g, \tau)
\end{equation}
and for all $(\x,\y) \in \cX\times\cY$
\begin{align}
\LSE_{g^\star}(\x) &= \tau^\star(\x)  \\
p_{g^\star}(\y|\x) &= q(\y|\x) \exp(g^\star(\x,\y) - \tau^\star(\x)).
\end{align}
\end{proposition}
See Appendix \ref{proof:min_min_mle} for a proof. The optimality in Proposition \ref{prop:min_min_mle} holds because we perform minimization in the space of continuous functions. If we parameterize $\tau$ as a neural network, and perform minimization in the space of parameters instead,
our approach only performs approximate MLE. However, thanks to the universality of neural networks, minimization in the space of parameters should be close to minimization in the space of continuous functions, provided that the neural network used is sufficiently expressive. This is confirmed in Figure \ref{fig:learning_curves_cal500}. The larger the number of $\y'$ samples we draw to estimate the expectation in \eqref{eq:L_g_tau_expectation}, which corresponds to $B'$ in Algorithm \ref{algo:doubly_stochastic}, the faster we converge to the exact MLE objective.

\paragraph{Generalization ability of the learned log-partition.}

Our approach not only provides a tractable objective criterion to learn EBMs by SGD, but also a novel means to estimate the log-partition function on unseen data points. Indeed, because we parameterize $\tau_\v$ as a neural network, we can evaluate it on new datapoints at inference time. Figure \ref{fig:tau} empirically confirms the generalization ability of our learned log-partition function on unseen data points.

Learning the partition function, as opposed to computing it or estimating it, bears some similarity with learning the value function in reinforcement learning \citep{konda1999actor,schulman2017proximal}.
Parameterizing dual variables as neural networks, as we do in this work, has also been explored in optimal transport
\citep{seguy2017large, korotin2021neural} and in generative adversarial networks
\citep{nowozin2016f}.

\subsection{Finite sum setting}

The MLE objective \eqref{eq:mle} corresponds to the \textit{expected} risk minization setting, in which there are potentially infinitely-many $(\x, \y)$ pairs.
In practice, we often work in the \textit{empirical} risk minimization or finite sum setting, in which we  use a finite number of pairs $(\x_1,\y_1)$, ..., $(\x_n, \y_n)$.
The classical MLE objective \eqref{eq:mle} then becomes (removing constants)
\begin{equation}
\label{eq:mle_finite_sum}
\widetilde{\cL}_{\MLE}(g) 
\coloneqq \frac{1}{n} \left(\sum_{i=1}^n \LSE_g(\x_i) - g(\x_i, \y_i)\right)
\end{equation}
and our proposed objective \eqref{eq:proposed_mle} becomes
\begin{equation}
\label{eq:min_min_finite_sum}
\widetilde{\cL}_{\MLE}(g, \tau) 
\coloneqq \frac{1}{n} \left(\sum_{i=1}^n L_{g,\tau}(\x_i) - g(\x_i, \y_i)\right).
\end{equation}
In the finite sum setting, if we do not need $\tau$ to generalize to unseen $\x$ instances (recall that the log-partition is not needed at inference time if all we want to compute is the mode, i.e., the argmax prediction), then we can set
$\tau_\v(\x_i) \coloneqq v_i$, 
for $\v \coloneqq (v_1, \dots, v_n) \in \RR^n$.
Our approach then solves the finite-sum MLE objective optimally.
\begin{proposition}{Optimality, finite sum setting}\label{prop:finite}

Suppose $\tau_\v(\x_i) \coloneqq v_i$ for $i \in [n]$. Then,
\begin{equation}
\min_{\w \in \cW} \widetilde{\cL}_{\MLE}(g_\w)
=
\min_{\substack{\w \in \cW\\ \v \in \RR^n}} \widetilde{\cL}_{\MLE}(g_\w, \tau_\v)
\end{equation}
and $v^\star_i = \LSE_{g_{\w^\star}}(\x_i)$ for all $i \in [n]$.
\end{proposition}
See Appendix \ref{proof:finite} for a proof. Remarkably, the result holds \textit{even if} $g_\w$ is nonlinear in $\w \in \cW$.

\paragraph{Joint convexity in the case of linear models.}

If $g_\w(\x_i, \y_i) = \langle h_\w(\x_i), \y_i \rangle$, where $h_\w$ is linear in $\w$,
then the classical MLE objective \eqref{eq:mle_finite_sum} is that of conditional random fields \citep{lafferty2001conditional,sutton2012introduction} and it is convex in $\w$. However, despite convexity, it could be intractable if for example $\cY$ is the set of permutations. 
In contrast, our objective \eqref{eq:min_min_finite_sum}
is jointly convex in $(\w, \v)$, as proved in Appendix \ref{proof:joint_convexity}, and it remains tractable as long as we can sample from $q(\cdot|\x)$.
In this convex setting, SGD enjoys a convergence rate of $O(\nicefrac{1}{\sqrt{t}})$, and if we further add strongly-convex regularization on the parameters, the rate becomes $O(\nicefrac{1}{t})$ \cite{garrigos2023handbook,bach2024learning}.
Therefore, our objective can be optimized using stochastic gradient methods, and we have obtained convergence rates for learning EBMs in arbitrary combinatorial spaces.

\subsection{Comparison with min-max formulation}\label{subsec:minmax}

One distinguishing feature of our approach is that it uses a min-min formulation. 
It is therefore insightful to compare it with a min-max formulation.
With some overloading of the notation, let us define
\begin{align}
\cL_\MLE(g, p)
\coloneqq
&~ \EE_\x 
\EE_{\y' \sim p(\cdot|\x)}[g(\x, \y')] - \mathrm{KL}(p(\cdot|\x), q(\cdot|\x)) \\
&- \EE_{(\x,\y)}[g(\x, \y)] + \mathrm{const}.
\end{align}
We then have the following proposition.
\begin{proposition}{Optimality of min-max, MLE case}
\label{prop:min_max_mle}

We have
\begin{equation}
\min_{g \in \cF(\cX\times\cY)} \cL_{\MLE}(g)
=\min_{g \in \cF(\cX\times\cY)}
\max_{p \in \cP_1(\cY|\cX)} \cL_{\MLE}(g, p).
\end{equation}
\end{proposition}
A proof is given in Appendix \ref{proof:min_max_mle}.
This is akin to adversarial approaches,
where the function $g$ plays the role of a discriminator and the distribution $p$ plays the role of a generator \citep{nowozin2016f,ho2016generative}.
Because the maximization \wrt $p$ in the space of distributions is usually intractable when $\cY$ is combinatorial or infinite
and because we need to be able to sample from $p(\cdot|\x)$,
it is common to replace $p$ with a parameterized distribution from which it is easy to sample. 
An additional challenge comes from gradient computations: since $p$ appears in the sampling, one typically uses the score function estimator (REINFORCE) to estimate the gradient w.r.t. the parameters of $p$. However, this estimator suffers from high variance. In contrast, in our min-min approach, the variable $\tau$ is a function, not a distribution and its gradients are easy to compute, since expectations are \wrt $q$,
not $p$.

\begin{figure*}[t]
    \centering
    \subfigure[yeast]{\includegraphics[width=0.32\textwidth]{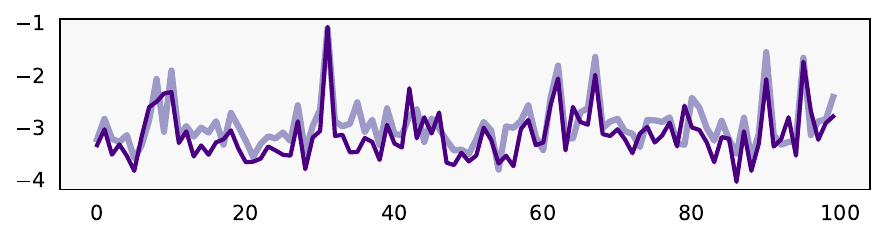}} 
    \subfigure[scene]{\includegraphics[width=0.32\textwidth]{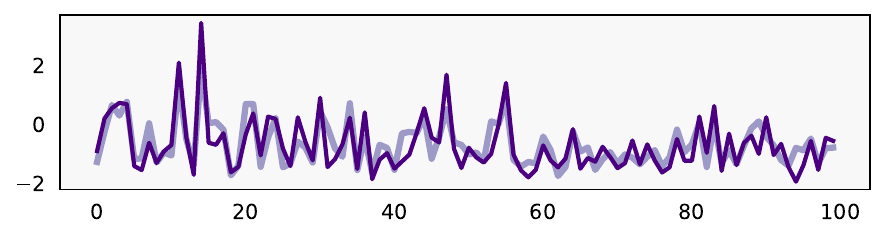}} 
    \subfigure[birds]{\includegraphics[width=0.32\textwidth]{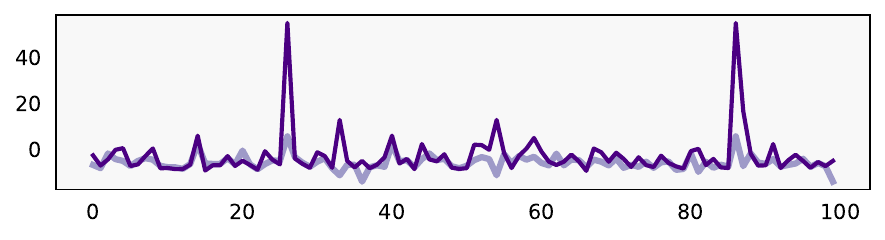}} \\
     \vspace{-.5em}
    \subfigure[emotions]{\includegraphics[width=0.32\textwidth]{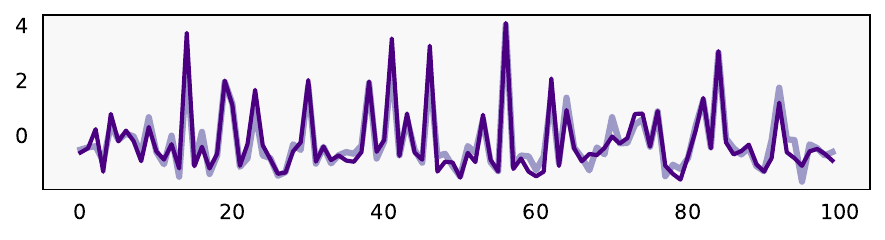}} 
    \subfigure[cal500]{\includegraphics[width=0.32\textwidth]{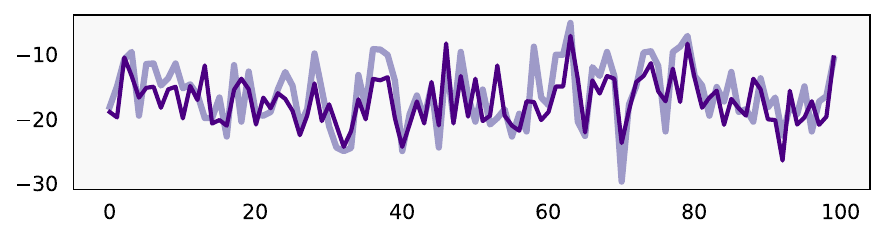}} 
     \vspace{-.5em}
    \caption{ \footnotesize{\bf Generalization ability of the learned log-partition function for multilabel classification.}
    As a testbed to compare the learned log-partition (dark purple),
    we use the unary model (see Section \ref{sec:multilabel} for details), which enjoys a closed-form expression for the exact log-partition (light purple).
    We pick randomly $100$ test samples ($x$-axis) on 5 multilabel classification datasets, after training models with hyper-parameters optimized against the validation set.
    }
    \label{fig:tau}
\end{figure*}

\subsection{Generalization to Fenchel-Young losses}

Another advantage of our proposed approach is that it naturally generalizes to any loss function in the Fenchel-Young family \citep{blondel2020learning}. 
It is well-known that the log-sum-exp can be interpreted as a ``softmax''. To obtain a more general notion of softmax,
we replace the KL divergence in \eqref{eq:lse_variational} with $f$-divergences \citep{csiszar1967information,ali1966general}, which are defined when $\cY$ is a finite set as
\begin{equation}
D_f(p, q) \coloneqq \sum_{\y \in \cY} f(p(\y)/q(\y)) q(\y).
\end{equation}
We assume that $f$ is strictly convex and differentiable on $(0, +\infty)$.
We can then define the $f$-softmax as
\begin{equation}
\softmax_g^f(\x) \coloneqq \max_{p \in \cP_1(\cY)} \langle g(\x, \cdot), p \rangle - D_f(p, q(\cdot|\x)),
\end{equation}
and the $f$-softargmax as
\begin{equation}
p_g^f(\cdot|\x) \coloneqq \argmax_{p \in \cP_1(\cY)} \langle g(\x, \cdot), p \rangle - D_f(p, q(\cdot|\x)).
\end{equation}

By analogy with the MLE objective in \eqref{eq:mle},
we can then consider the objective
\begin{equation}
\cL_f(g) 
\coloneqq \EE_\x\left[\softmax_g^f(\x)\right]
- \EE_{(\x,\y)} \left[g(\x, \y)\right].
\end{equation}
Unfortunately, just as for the log-sum-exp in the MLE setting, 
the $f$-softmax is intractable to compute when $\cY$ is combinatorially large.
Similarly to \eqref{eq:proposed_mle}, using duality arguments, we arrive at the objective
\begin{equation}
\cL_f(g, \tau) 
\coloneqq \EE_\x\left[L_{g,\tau}^f(\x)\right]
- \EE_{(\x,\y)} \left[g(\x, \y)\right],
\end{equation}
with
$
L_{g,\tau}^f(\x)
\coloneqq
\tau(\x) + \sum_{\y \in \cY} q(\y|\x) f_+^*(g(\x, \y) - \tau(\x)).
$

Here, $f_+^*$ is the convex conjugate of 
$f_+$, the restriction of $f$ to $\RR_+$.
We have the next generalization of Proposition \ref{prop:min_min_mle}.
\begin{proposition}{Optimality of min-min, general case}
\label{prop:min_min_fy}

Let $f$ be a strictly convex differentiable function such that $(0, +\infty) \subseteq \dom f'$. Suppose that for all $\y \in \cY$, $\x \mapsto q(\y|\x)$ is continuous. 
Then, we have
\begin{equation}
\min_{g \in \cF(\cX\times\cY)} \cL_f(g)
=\min_{\substack{g \in \cF(\cX\times\cY)\\\tau \in \cF(\cX)}} \cL_f(g, \tau)
\end{equation}
and for $(\x,\y) \in \cX\times\cY$
\begin{align}
p_{g^\star}^f(\y|\x) &= q(\y|\x) (f_+^*)'(g^\star(\x, \y) - \tau^\star(\x)).
\end{align}
\end{proposition}
A proof is given in Appendix \ref{proof:min_min_fy}.
In the MLE case, $\tau$ coincided exactly with the log-partition function.
In the more general setting, $\tau$ corresponds to the Lagrange multiplier associated with the equality constraints
$\sum_{\y \in \cY} p_g^f(\y|\x) = 1$.

\paragraph{Example: sparsemax for EBMs.}

When $f(u) = \frac{1}{2}(u^2 - 1)$, which is the generating function of the chi-square divergence, 
we obtain the first tractable method for optimizing the sparsemax loss \citep{martins2016softmax} on EBMs in combinatorially-large discrete spaces. We have
\begin{equation}
L_{g,\tau}^f(\x)
=
\tau(\x) + \frac{1}{2} \sum_{\y \in \cY} q(\y|\x) \left[g(\x, \y) - \tau(\x)\right]_+^2
\end{equation}
and
$p_{g^\star}^f(\y|\x) = q(\y|\x) \left[g^\star(\x,\y) - \tau^\star(\x)\right]_+,$
where $[\cdot]_+  \coloneqq \max(., 0)$ is the non-negative part.
Previous attempts to use the sparsemax loss for structured prediction require a $k$-best maximization oracle \citep{pillutla2018smoother}. Unfortunately, such oracles are not available for all sets $\cY$ (e.g., permutations) and their cost is usually polynomial in $k$, while $k$ could be arbitrarily large. In contrast, our doubly-stochastic approach only requires us to be able to sample from the reference probability measure $q(\cdot|\x)$.

\begin{table*}[ht]
\scriptsize
\centering
\caption{$f_1$-score on multi-label classification for different models and losses (with constant $\tau$ model, that is $\tau_\v(\x_i) \coloneqq v_i$.)}
\begin{tabular}{cccccccc }
\toprule
Loss & Model $g$ & yeast & scene & birds & emotions & cal500 \\
\midrule
Logistic (min-min) & Linear (unary) & 63.41 & 70.78 & 45.49 & 61.63 & 40.19 \\
 & MLP (unary) & 65.04 & 75.35 & \textbf{46.48} & 65.33 & 44.91 \\
 & ResNet (unary) & 65.03 & 75.64 & 36.93 & \textbf{67.45} & 45.23 \\
 & Linear (pairwise) & 63.40 & 70.72 & 45.49 & 61.63 & 40.23 \\
 & MLP (pairwise) & \textbf{65.11} & 75.22 & 46.08 & 64.84 & 45.00\\
 & ResNet (pairwise) & 64.84 & 75.68 & 41.34 & 64.26 & 44.61 \\
Sparsemax (min-min) & Linear (unary) & 62.92 & 70.43 & 45.12 & 63.14 & 45.18 \\
 & MLP (unary) & 63.22 & 75.83 & 42.42 & 66.86 & 46.39 \\
 & ResNet (unary) & 62.27 & 75.44 & 28.63 & 44.53 & 39.11 \\
 & Linear (pairwise) & 62.91 & 69.89 & 45.12 & 63.14 & 45.18 \\
 & MLP (pairwise) & 63.39 & 74.94 & 40.68 & 66.28 & 46.39\\
 & ResNet (pairwise) & 62.32 & \textbf{75.63} & 27.85 & 64.58 & 30.93 \\
 \midrule
  \midrule
Logistic (exact MLE) & Linear (unary) & 61.46 & 70.64 & 43.97 & 63.35 & 37.31 \\
 & MLP (unary) & 60.71 & 70.81 & 40.88 & 62.99 & 36.24 \\
 & Resnet (unary) & 62.27 & 71.02 & 40.78 & 61.50 & 37.45 \\
 \midrule
Generalized Fenchel-Young & Linear (unary) & 61.41 & 70.26 & 44.13 & 61.70 & 36.85 \\
 & MLP (unary) & 61.31 & 70.84 & 40.73 & 61.85 & 37.35 \\
 & Linear (pairwise) & 61.62 & 70.42 & 42.28 & 63.05 & 35.12 \\
 \midrule
Logistic (MCMC) & Linear (unary) & 61.98 & 70.53 & 44.93 & 61.76 & 44.61 \\
 & MLP (unary) & 60.34 & 71.57 & 33.13 & 62.23 & 45.70 \\
 & Resnet (unary) & 56.57 & 71.24 & 27.03 & 58.98 & 44.34 \\
 & Linear (pairwise) & 61.96 & 70.50 & 44.93 & 61.76 & 44.61 \\
 & MLP (pairwise) & 60.83 & 71.92 & 34.61 & 63.15 & 45.95 \\
 & Resnet (pairwise) & 56.68 & 70.46 & 27.08 & 64.05 & 44.35 \\
\midrule
Logistic (min-max) & Linear (unary) & 52.80 & 52.29 & 27.59 & 61.72 & 35.08 \\
 & MLP (unary) & 58.58 & 52.54 & 24.95 & 60.53 & 46.08 \\
 & Resnet (unary) & 57.53 & 42.82 & 25.32 & 55.85 & 47.59 \\
 & Linear (pairwise) & 52.80 & 52.29 & 27.59 & 61.72 & 35.08 \\
 & MLP (pairwise) & 58.76 & 51.59 & 26.71 & 58.68 & 43.31 \\
 & Resnet (pairwise) & 57.53 & 42.95 & 22.35 & 46.72 & \textbf{47.67} \\
\bottomrule
\end{tabular}\label{tab:f1_losses}
\end{table*}
\section{Experiments}
\label{sec:experiments}

We now demonstrate our approach through experiments on multilabel classification and label ranking. 
Throughout our experiments, as in \eqref{eq:g_decomposition},
we assume that
$g(\x, \y) = \Phi(h(\x), \y)$
where $\Phi(\thetav, \y)$ is a coupling function and $h(\x)$ is a model function producing logits $\thetav$. 
Experimental details and additional results are presented in Appendix \ref{app:exp_details}.

\subsection{Multilabel classification}
\label{sec:multilabel}

In this section, we perform experiments on multilabel classification with $k$ classes.
Here, given an input $\x$, the goal is to predict an output $\y$ 
belonging to the powerset of $[k]$, which can be represented as $\cY = \{0,1\}^k$. 
The cardinality is $|\cY| = 2^k$ and the convex hull is $\conv(\cY) = [0,1]^k$.

\paragraph{Unary model.}

To perform multilabel classification, the simplest approach is to use the bilinear coupling \eqref{eq:bilinear_coupling} $\Phi(\thetav, \y) = \sum_{j=1}^k \theta_j y_j$ together with logits $\thetav = h(\x) \in \RR^k$. In this case, predicting the mode \eqref{eq:argmax} with uniform $q$ reads 
\begin{equation}
\y^\star_g(\x)
= \argmax_{\y \in \{0,1\}^k} \Phi(\thetav, \y)
=
\argmax_{\y \in \{0,1\}^k} \langle \thetav, \y \rangle,
\end{equation}
for which the optimal solution is
\begin{equation}
[\y^\star_g(\x)]_j =
\begin{cases}
1 &\mbox{ if } \theta_j \ge 0 \\
0 &\mbox{ if } \theta_j < 0
\end{cases}
\quad \forall j \in [k].
\end{equation}
The log-sum-exp \eqref{eq:lse} and marginal inference \eqref{eq:conditional_expectation} both enjoy closed-form solutions:
\begin{align}
\LSE_g(\x) &= \sum_{j=1}^k \softplus(\theta_j) \label{eq:lse_unary_model} \\
[\muv_g(\x)]_j &= \sigmoid(\theta_j) \quad \forall j \in [k].
\end{align}
Therefore, the unary model boils down to placing sigmoid activations on top of the logits $\thetav = h(\x)$, as usual in deep learning. While our approach is not needed in the unary model, the unary model is a useful testbed for our method,
as we can use \eqref{eq:lse_unary_model} as a ground-truth for evaluating $\tau$.

\begin{table*}[ht]
 \scriptsize
\centering
\caption{Kendall rank correlation coefficient on label ranking for different models and losses, with MLP $\tau$ model}
\begin{tabular}{ccccccccc}
\toprule
Loss & Polytope $\cC$ & Model $g$ & authorship & glass & iris & vehicle & vowel & wine \\
\midrule
Logistic (min-min) & $\cP$ & Linear & 85.14 & 80.47 & 57.78 & 79.74 & 50.82 & 91.36 \\
 & $\cP$ & MLP & 88.69 & 84.39 & 97.78 & 86.27 & \textbf{74.26} & 90.12 \\
 & $\cB$ & Linear & 87.71 & 81.29 & 97.78 & 83.33 & 56.33 & 95.06 \\
 & $\cB$ & MLP & \textbf{90.80} & 85.43 & 98.52 & 86.41 & 65.20 & 91.98 \\
 \midrule
Sparsemax (min-min) & $\cP$ & Linear & 84.88 & 79.84 & 52.59 & 78.76 & 50.25 & 93.83 \\
 & $\cP$ & MLP & 86.52 & \textbf{85.94} & \textbf{99.26} & 85.75 & 68.34 & 91.36 \\
 & $\cB$ & Linear & 89.09 & 79.43 & 97.78 & 82.81 & 56.24 & 90.12 \\
 & $\cB$ & MLP & 87.84 & 80.47 & 98.52 & \textbf{87.19} & 55.43 & 85.80 \\
  \midrule
   \midrule
 Logistic (MCMC) & $\cP$ & Linear & 84.29 & 80.67 & 55.56 & 79.15 & 49.97 & 91.36 \\
 & $\cP$ & MLP & 87.90 & 75.50 & 97.04 & 86.34 & 57.90 & 91.98 \\
  & $\cB$ & Linear & 88.10 & 79.33 & 97.78 & 82.22 & 55.99 & \textbf{96.91} \\
 & $\cB$ & MLP & 89.68 & 80.16 & 97.78 & 85.88 & 48.90 & 91.98  \\
  \midrule
 Logistic (min-max) & $\cP$ & Linear & 61.01 & 79.22 & 61.48 & 70.20 & 45.68 & 65.43 \\
 & $\cP$ & MLP & 70.02 & 67.03 & 55.56 & 68.56 & 40.22 & 70.99 \\
  & $\cB$ & Linear & 82.58 & 78.40 & 61.48 & 74.90 & 52.86 & 93.21 \\
 & $\cB$ & MLP & 78.50 & 71.68 & 42.96 & 68.76 & 48.99 & 83.33 \\
\bottomrule
\end{tabular}
\label{tab:kendall_loss_main}
\end{table*}
\paragraph{Pairwise model (Ising model).}

To try our approach on a model for which a closed form is not available for the log-partition, we consider a linear-quadratic coupling,
\begin{align}
\Phi(\thetav, \y) 
&\coloneqq \langle \u, \y \rangle 
+ \frac{1}{2} \langle \y, \U \y \rangle \\
&= \sum_{j=1}^k u_j y_j + \frac{1}{2} \sum_{i=1}^k \sum_{j=1}^k U_{i,j} y_i y_j,   
\end{align}
where $\thetav \coloneqq (\u, \U) = h(\x) \in \RR^k \times \RR^{k \times k}$. 
As its name indicates, this coupling is linear in $\thetav$, but quadratic in $\y$.
The model function $h$ associates weights $u_j$ to labels $y_j$ and weights $U_{i,j}$ to pairwise label interactions $y_i y_j$.
The relaxed inference problem \eqref{eq:relaxed_argmax} becomes
\begin{equation}
\y^\star_g(\x) 
\approx
\argmax_{\muv \in [0,1]^k} \Phi(\thetav, \muv)
=
\argmax_{\muv \in [0,1]^k} \langle \u, \muv \rangle 
+ \frac{1}{2} \langle \muv, \U \muv \rangle.
\end{equation}
Contrary to the unary model case, there is no longer a closed-form solution.
However, if we assume that $\U$ is negative semi-definite (which is easy to impose),
this problem is concave in $\muv$ and can be solved optimally without learning rate tuning using coordinate ascent
\citep{gfy}.

\paragraph{Results.}

We evaluate our models on classical multilabel classification datasets.  The
hyperparameters (learning rate and regularization) are optimized on a validation
set. Once the best hyperparameters are found, we refit the model on the combined
training and validation sets and evaluate it on the test set. Our results in
Table \ref{tab:f1_losses} show that the logistic and sparsemax losses trained
with our approach work better than the generalized Fenchel-Young loss as
well as min-max and MCMC sampling approaches in various configurations.  
For the min-max approach, we use optimistic ADAM as
solver, an MLP as generator and we use REINFORCE (score function estimator) for
gradient estimation. For MCMC sampling, we use standard Metropolis–Hastings
algorithm with uniform proposal distribution. We also present learning curves
in Figure \ref{fig:learning_curves_cal500}. We numerically validate that our
approach converges to true MLE as more $\y'$ samples are used. We also observe a
regularization phenomenon: using fewer $\y'$ samples leads to better $f_1$
scores than exact MLE. These findings are confirmed on other datasets (Figure
\ref{fig:learning_curves_all}). 

\paragraph{Generalization ability of the learned partition function.}

A natural question is whether the learned log-partition $\tau$ generalizes to unseen test samples, that is, whether it correctly approximates the true log partition function. To empirically demonstrate this, we consider the unary model, for which the log-sum-exp enjoys the closed form solution \eqref{eq:lse_unary_model}. We parameterize $\tau$ as an MLP and $g$ as a linear network. After training, we evaluate the learned $\tau$ on $100$ randomly sampled test samples and compare it with the true log-sum-exp. Results are shown in Figure \ref{fig:tau}. We observe strong correlation between the learned and true log-sum-exp.

\subsection{Label ranking}

In this section, we perform label ranking of $k$ labels.
Given an input $\x$, the goal is to predict a ranking $\y$ of the labels, that is, a permutation of $(1, \dots, k)$. The cardinality is then $|\cY| = k!$.
We focus on the bilinear coupling $\Phi(\thetav, \y) = \langle \thetav, \y \rangle$.
We consider two representations for permutations.

\paragraph{Permutahedron.}

A permutation can be represented as a vector.
In this case, $\conv(\cY) = \cP$, where $\cP$ is the permutahedron \citep{bowman1972permutation,ziegler2012lectures, blondel2020fast}.
We set the logits to $\thetav = h(\x) \in \RR^k$.
The argmax
$
\y^\star_g(\x)
= \argmax_{\y \in \cY} \Phi(\thetav, \y)
=
\argmax_{\y \in \cP} \langle \thetav, \y \rangle
$
can be solved optimally using an argsort \citep[Proposition 1]{sander2023fast}. 
However, $\LSE_g$ and $\muv_g$ are known to be \#P-complete to compute,
making classical MLE intractable.

\paragraph{Birkhoff polytope.}

A permutation can also be represented as a permutation matrix.
In this case, $\conv(\cY) = \cB$, where $\cB$ is the Birkhoff polytope, the set of doubly-stochastic matrices.
The logits are $\thetav = h(\x) \in \RR^{k \times k}$.
The argmax
$
\y^\star_g(\x)
= \argmax_{\y \in \cY} \Phi(\thetav, \y)
=
\argmax_{\y \in \cB} \langle \thetav, \y \rangle
$
can be solved optimally using the Hungarian algorithm \citep{kuhn1955hungarian}. 
However, $\LSE_g$ and $\mu_g$ are again \#P-complete to compute,
making classical MLE intractable.
\vspace{-1em}
\paragraph{Results.}

We use the same procedure as before to tune the hyperparameters.  Our main results are in Table \ref{tab:kendall_loss_main}, confirming that our approach successfully learns in the space of permutations.
The Birkhoff polytope works better than the permutahedron when using linear models. This makes sense because the logits when using the Birkhoff polytope are in $\RR^{k \times k}$, enabling to capture label interactions even with a linear model, while the logits when using the permutahedron are in $\RR^k$, which gives the ability to assign weight to single labels only. However, when using an MLP, we observe that the permutahedron works as well as the Birkhoff polytope, which is thanks to nonlinearity. The logistic and the sparsemax losses perform comparably overall.

\section{Conclusion and future work}

In this paper, we proposed a novel min-min formulation for jointly learning probabilistic EBMs and their log-partition in combinatorially-large spaces. 
Our method provably solves the MLE objective when minimizing the expected risk in the space of continuous functions.
To optimize the model parameters, we proposed a doubly-stochastic MCMC-free SGD scheme, which only requires the ability to sample outputs from a prior reference probability distribution. We experimentally showed that parameterizing $\tau$ as a neural network leads to successful generalization on unseen data points. We demonstrated our method through experiments on multilabel and label ranking tasks.

In this paper, we evaluated our proposed method primarily in the prediction
setting (finding the mode). Our paper does not directly address the challenge of
sampling from $p(\y|\x)$, which is central to evaluating performance in a
probabilistic or generative setting. We leave to future work whether the learned
log-partition function can be used to design better samplers.

\section*{Acknowledgements}

We thank Alexandre Ram\'{e} as well as the anonymous reviewers for constructive feedbacks on this paper.

\section*{Impact statement}

This paper introduces a novel formulation for learning conditional EBMs. We do not foresee any specific ethical or societal implications arising directly from this work.

\bibliography{references}
\bibliographystyle{icml2025}

\newpage
\appendix
\onecolumn

\section{Experimental details and additional results}\label{app:exp_details}

\begin{figure}[H]
    \centering
    \begin{tabular}{cccc}
        & \text{Loss (train)} & \text{Gradient norms (train)} & \text{$f_1$-score (test)} \\
    \raisebox{2.5\height}{\rotatebox[origin=c]{90}{Yeast}} & \includegraphics[width=0.3\textwidth]{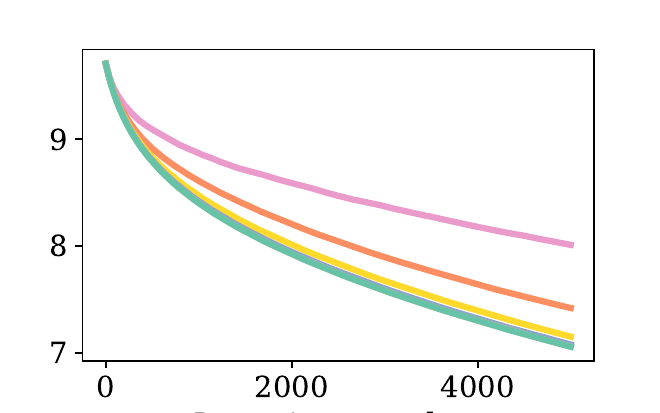} & \includegraphics[width=0.3\textwidth]{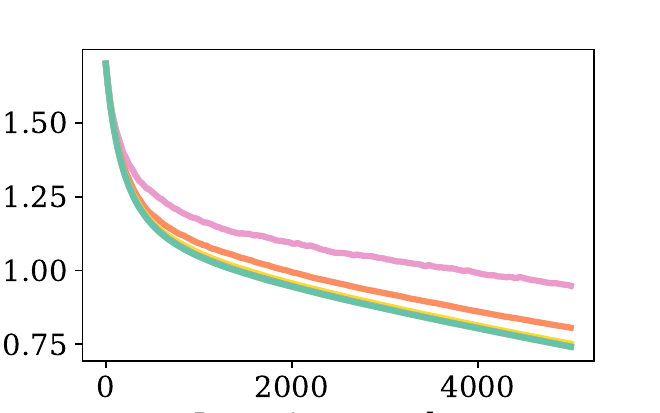} & \includegraphics[width=0.3\textwidth]{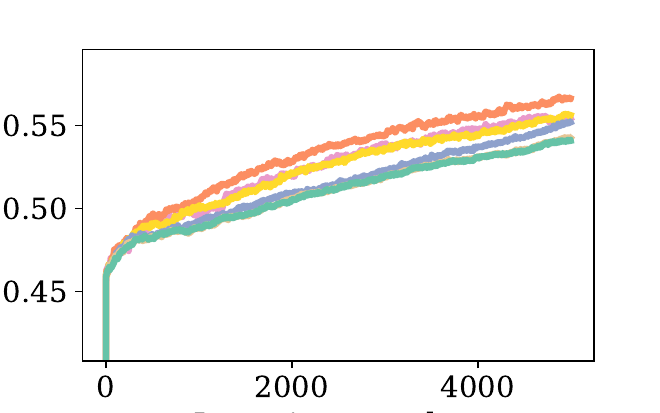} \\
    \raisebox{2.5\height}{\rotatebox[origin=c]{90}{Scene}} & \includegraphics[width=0.3\textwidth]{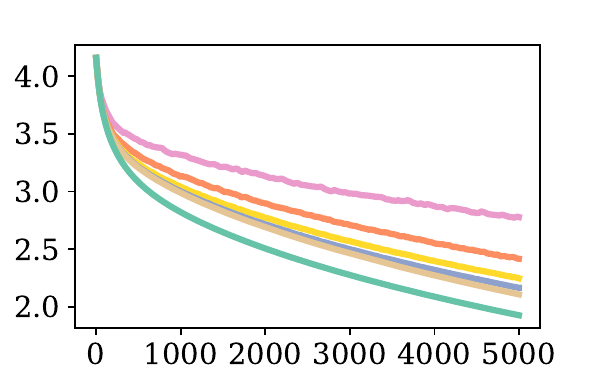} & \includegraphics[width=0.3\textwidth]{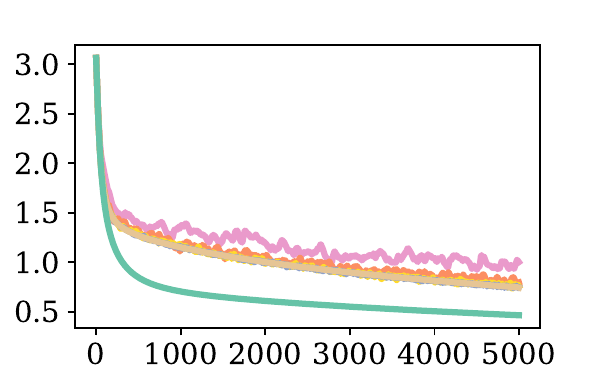} & \includegraphics[width=0.3\textwidth]{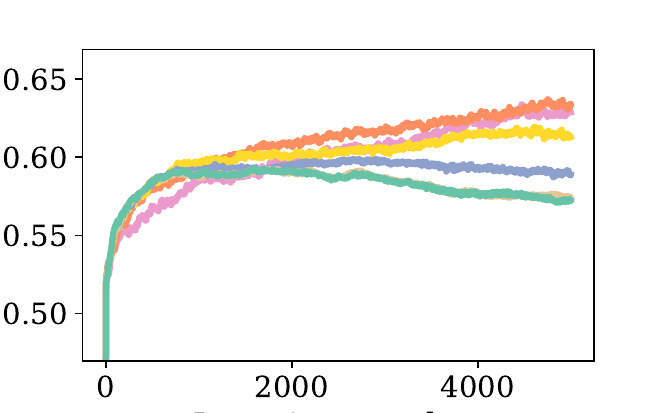} \\
    \raisebox{2.\height}{\rotatebox[origin=c]{90}{Emotions}} & \includegraphics[width=0.3\textwidth]{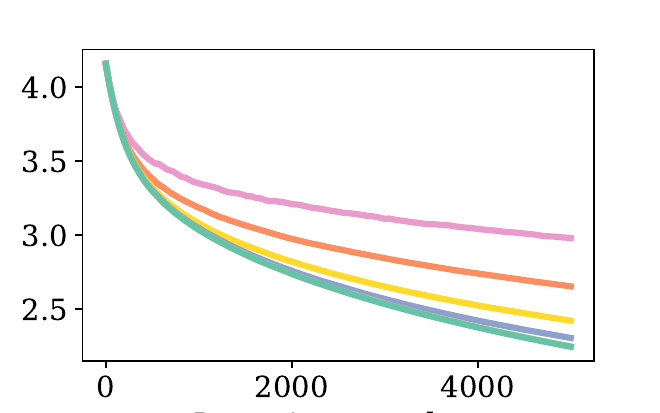} & \includegraphics[width=0.3\textwidth]{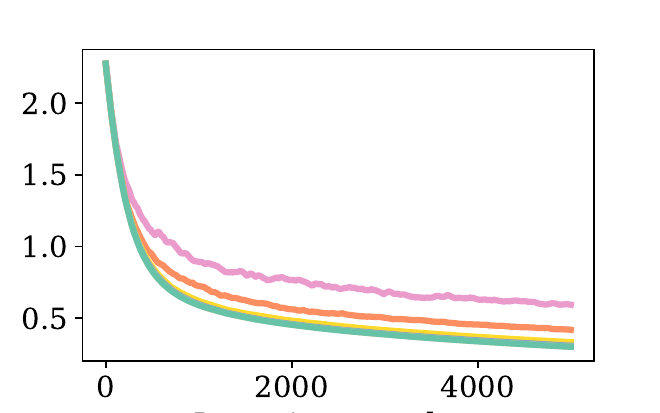} & \includegraphics[width=0.3\textwidth]{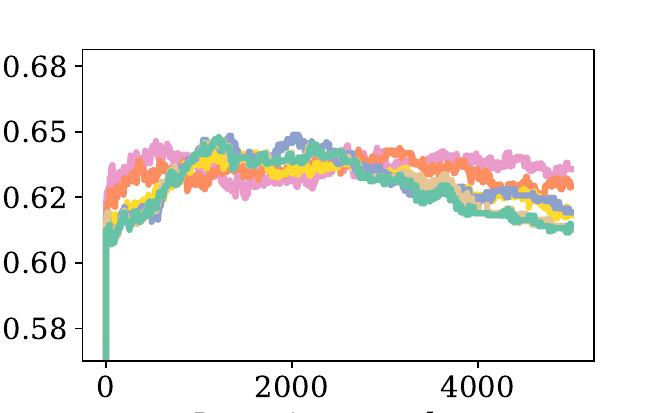} \\
    \raisebox{2.5\height}{\rotatebox[origin=c]{90}{Birds}} & \includegraphics[width=0.3\textwidth]{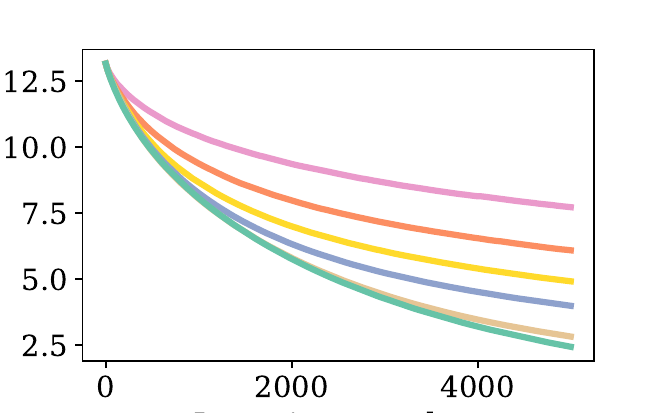} & \includegraphics[width=0.3\textwidth]{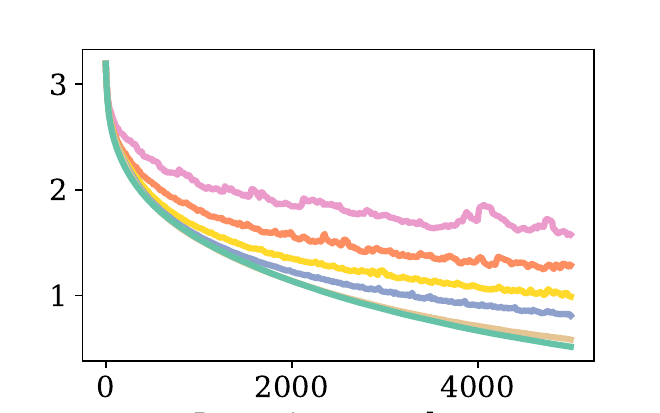} & \includegraphics[width=0.3\textwidth]{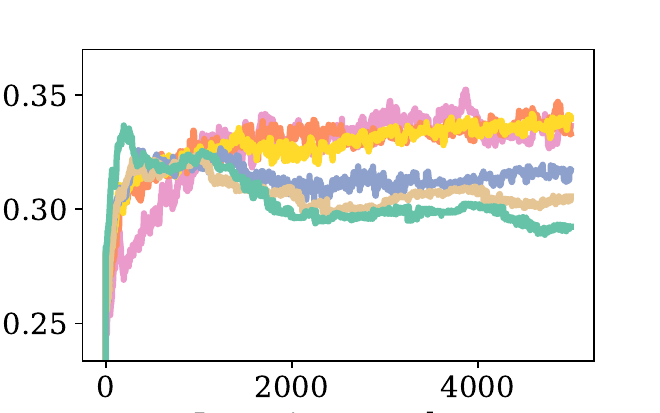} \\
     \raisebox{2.\height}{\rotatebox[origin=c]{90}{Mediamill}} & \includegraphics[width=0.3\textwidth]{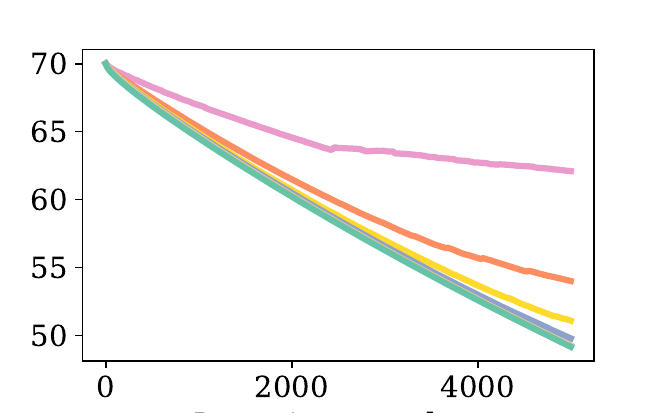} & \includegraphics[width=0.3\textwidth]{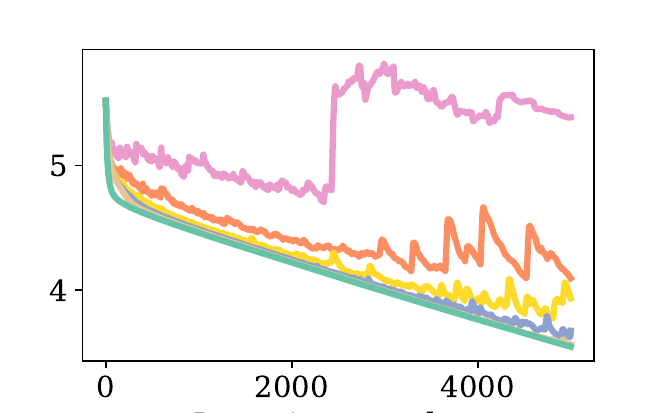} & \includegraphics[width=0.3\textwidth]{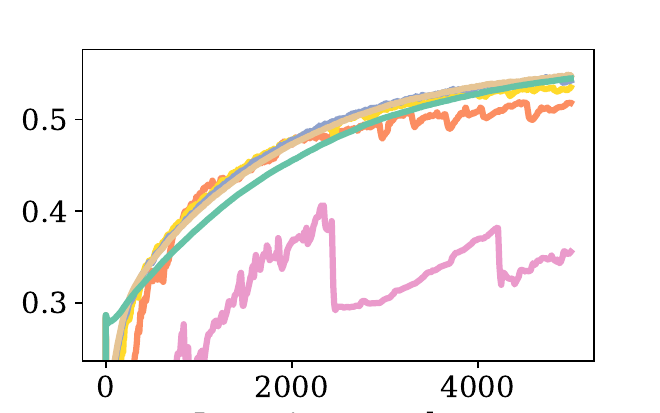}
    \end{tabular}
    \includegraphics[width=0.6\textwidth]{figures/legend.pdf} \\
    \caption{{\bf Convergence of the proposed approach as a function of the number of prior samples $\y'$ drawn.} 
    To be able to compute the exact MLE objective \eqref{eq:mle_finite_sum}, we use the unary multilabel model (Section \ref{sec:multilabel}) as the test bed. Note that the loss and gradient in the left and center columns are computed using \eqref{eq:mle_finite_sum} even for our method. We make two key observations: i) our approach converges to exact MLE as predicted by our theory, ii) the number of $\y'$ samples can have a regularization effect on the test $f_1$-score.}\label{fig:learning_curves_all}
\end{figure}

Our implementation is made using JAX \citep{jax2018github}.

\paragraph{Multilabel classification datasets.}

We use the same datasets as in \citet{gfy}.
The datasets can be downloaded from
\url{https://www.csie.ntu.edu.tw/~cjlin/libsvmtools/datasets/}.
The dataset characteristics are described in Table \ref{tab:dataset_characteristics} below.

\begin{table}[h]
\centering
\caption{Dataset Characteristics} %
\label{tab:dataset_characteristics} %

\begin{tabular}{cccccccc}
\toprule
Dataset & Type & Train & Dev & Test & Features & Classes & Avg. labels \\
\midrule
Birds & Audio & 134 & 45 & 172 & 260 & 19 & 1.96 \\
Cal500 & Music & 376 & 126 & 101 & 68 & 174 & 25.98 \\
Emotions & Music & 293 & 98 & 202 & 72 & 6 & 1.82 \\
Mediamill & Video & 22,353 & 7,451 & 12,373 & 120 & 101 & 4.54 \\
Scene & Images & 908 & 303 & 1,196 & 294 & 6 & 1.06 \\
Yeast & Micro-array & 1,125 & 375 & 917 & 103 & 14 & 4.17 \\
\bottomrule
\end{tabular}

\end{table}

\paragraph{Convergence curves.}

Convergence curves are in Figure \ref{fig:learning_curves_all}. We use a linear model for $g$ (unary model), and an MLP for $\tau$ with ReLU acivation and a hidden dimension of $128$. Models are trained with the logistic loss. We use the Adam optimizer \citep{kingma2014adam} with a learning rate of $10^{-4}$ for the parameters of both $g$ and $\tau$. The models are trained for $5000$ steps with full batch \wrt $(\x_i, \y_i)$ pairs. The reason why we use full batches is to reduce the noise in the gradients and obtain a more stable training procedure.

\paragraph{Results over different models and losses.}

Additional experimental results are shown in Tables \ref{tab:f1_losses_tau_icnn} and \ref{tab:model}.

\begin{table}[H]
\centering
\caption{$f_1$-score on multi-label classification for different models and losses, with ICNN $\tau$ model}
\begin{tabular}{cccccccc}
\toprule
Loss & Model $g$ & yeast & scene & birds & emotions & cal500 \\
\midrule
logistic & Linear (unary) & 63.53 & 70.44 & 45.36 & 62.90 & 42.74 \\
 & MLP (unary) & 65.16 & 74.64 & \textbf{46.58} & 64.99 & 44.89 \\
 & Resnet (unary) & \textbf{65.40} & 75.60 & 38.71 & 67.05 & 45.23 \\
 & Linear (pairwise) & 63.53 & 70.47 & 45.36 & 62.90 & 42.75 \\
 & MLP (pairwise) & 64.95 & 74.84 & 44.11 & 64.52 & 45.00 \\
 & Resnet (pairwise) & 65.15 & 75.53 & 39.59 & 66.17 & 44.62 \\
sparsemax & Linear (unary) & 63.15 & 71.07 & 45.79 & 63.36 & 45.94 \\
 & MLP (unary) & 63.52 & 74.34 & 40.77 & \textbf{66.50} & 46.44 \\
 & Resnet (unary) & 62.62 & \textbf{76.79} & 30.18 & 63.38 & \textbf{46.64} \\
 & Linear (pairwise) & 63.15 & 71.05 & 45.79 & 63.36 & 45.94 \\
 & MLP (pairwise) & 63.25 & 74.93 & 41.61 & 66.18 & 46.40 \\
 & Resnet (pairwise) & 62.07 & 75.59 & 28.14 & 63.42 & 37.53 \\
\bottomrule
\end{tabular}\label{tab:f1_losses_tau_icnn}
\end{table}

\begin{table}[H]
\centering
\caption{$f_1$-score on multi-label classification for different models, with logistic loss}
\begin{tabular}{cccccccc}
\toprule
 Model $\tau$ & Model $g$ & yeast & scene & birds & emotions & cal500 \\
\midrule
Constant & Linear (unary) & 63.41 & 70.78 & \textbf{45.49} & 61.63 & 40.19\\
 & MLP (unary) & 65.04 & 75.35 & 46.48 & 65.33 & 44.91 \\
 & ResNet (unary) & 65.03 & \textbf{75.64} & 36.93 & \textbf{67.45} & 45.23 \\
 & Linear (pairwise) & 63.40 & 70.72 & \textbf{45.49} & 61.63 & 40.23 \\
 & MLP (pairwise) & 65.11 & 75.22 & 46.08 & 64.84 & 45.00 \\
 & ResNet (pairwise) & 64.84 & 75.68 & 41.34 & 64.26 & 44.61 \\
MLP & Linear (unary) & 61.51 & 70.77 & 44.65 & 60.93 & 42.40 \\
 & MLP (unary) & 62.03 & 70.81 & 39.70 & 62.26 & 46.24 \\
 & ResNet (unary) & 63.01 & 70.69 & 25.25 & 62.60 & 45.65 \\
 & Linear (pairwise) & 61.52 & 70.81 & 44.65 & 60.85 & 42.08 & \\
 & MLP (pairwise) & 62.42 & 69.96 & 38.89 & 63.38 & \textbf{46.08} \\
 & ResNet (pairwise) & 63.04 & 70.90 & 33.79 & 63.11 & 45.25 \\
ICNN & Linear (unary) & 63.53 & 70.44 & 45.36 & 62.90 & 42.42 \\
 & MLP (unary) & 65.15 & 74.65 & 44.99 & 64.99 & 44.89 \\
 & ResNet (unary) & \textbf{65.39} & 75.60 & 38.84 & 67.10 & 45.23 \\
 & Linear (pairwise) & 63.53 & 70.47 & 45.36 & 62.90 & 42.42\\
 & MLP (pairwise) & 64.94 & 74.84 & 45.82 & 64.52 & 45.00 \\
 & ResNet (pairwise) & 65.14 & 75.30 & 39.52 & 66.30 & 44.62 \\
\bottomrule
\end{tabular}\label{tab:model} 
\end{table}

\paragraph{Label ranking.}

The publicly-available datasets can be downloaded from
\url{https://github.com/akorba/Structured_Approach_Label_Ranking}. The experimental results are summarized in Tables \ref{tab:kendall_loss} and \ref{tab:polytope}. 

\begin{table}[H]
\centering
\caption{Kendall's tau on label ranking for different models and losses, with MLP $\tau$ model}
\begin{tabular}{ccccccccc}
\toprule
Loss & Polytope $\cC$ & Model $g$ & authorship & glass & iris & vehicle & vowel & wine \\
\midrule
logistic & $\cP$ & Linear & 85.14 & 80.47 & 57.78 & 79.74 & 50.82 & 91.36 \\
logistic & $\cP$ & MLP & 88.69 & 84.39 & 97.78 & 86.27 & \textbf{74.26} & 90.12 \\
logistic & $\cP$ & Resnet & 88.49 & 84.29 & 98.52 & \textbf{87.06} & 71.08 & 93.21 \\
logistic & $\cB$ & Linear & 87.71 & 81.29 & 97.78 & 83.33 & 56.33 & \textbf{95.06} \\
logistic & $\cB$ & MLP & \textbf{90.80} & 85.43 & 98.52 & 86.41 & 65.20 & 91.98 \\
logistic & $\cB$ & Resnet & 88.89 & 85.74 & 97.04 & 86.73 & 62.17 & 91.98 \\
sparsemax & $\cP$ & Linear & 84.88 & 79.84 & 52.59 & 78.76 & 50.25 & 93.83 \\
sparsemax & $\cP$ & MLP & 86.52 & \textbf{85.94} & \textbf{99.26} & 85.75 & 68.34 & 91.36 \\
sparsemax & $\cP$ & Resnet & 88.17 & 85.53 & 58.52 & 86.21 & 59.52 & 90.74 \\
sparsemax & $\cB$ & Linear & 89.09 & 79.43 & 97.78 & 82.81 & 56.24 & 90.12 \\
sparsemax & $\cB$ & MLP & 87.84 & 80.47 & 98.52 & 87.19 & 55.43 & 85.80 \\
sparsemax & $\cB$ & Resnet & 89.22 & 84.70 & 97.04 & 87.52 & 55.87 & 91.98 \\
\bottomrule
\end{tabular}\label{tab:kendall_loss}
\end{table}

\begin{table}[H]
\centering
\caption{Kendall's tau on label ranking for different models, with logistic loss}
\begin{tabular}{ccccccccc}
\toprule
Polytope $\cC$ & Model $\tau$ & Model $g$ & authorship & glass & iris & vehicle & vowel & wine \\
\midrule
$\cP$ & Constant & Linear & 84.94 & 79.64 & 57.78 & 79.61 & 49.18 & 91.98 \\
$\cP$ & Constant & MLP & 87.38 & 84.39 & 97.04 & 85.75 & 73.68 & 89.51 \\
$\cP$ & Constant & ResNet & 88.10 & 83.88 & \textbf{98.52} & \textbf{87.19} & 68.98 & 94.44 \\
$\cB$ & Constant & Linear & 88.17 & 77.16 & 97.78 & 83.66 & 57.29 & 95.06 \\
$\cB$ & Constant & MLP & 89.41 & 84.19 & \textbf{98.52} & 87.32 & 62.36 & 91.98 \\
$\cB$ & Constant & ResNet & 88.76 & 84.60 & \textbf{98.52} & 87.78 & 62.70 & 91.98 \\
 $\cP$ & MLP & Linear & 85.14 & 80.47 & 57.78 & 79.74 & 50.82 & 91.36 \\
 $\cP$ & MLP & MLP & 88.69 & 84.39 & 97.78 & 86.27 & \textbf{74.26} & 90.12 \\
 $\cP$ & MLP & Resnet & 88.49 & 84.29 & \textbf{98.52} & 87.06 & 71.08 & 93.21 \\
 $\cB$ & MLP & Linear & 87.71 & 81.29 & 97.78 & 83.33 & 56.33 & \textbf{95.06} \\
$\cB$ & MLP & MLP & \textbf{90.80} & 85.43 & \textbf{98.52} & 86.41 & 65.20 & 91.98 \\
$\cB$ & MLP & Resnet & 88.89 & \textbf{85.74} & 97.04 & 86.73 & 62.17 & 91.98 \\
$\cP$ & ICNN & Linear & 84.35 & 75.30 & 56.30 & 79.74 & 50.84 & 91.36 \\
$\cP$ & ICNN & MLP & 84.62 & 83.67 & 82.96 & 84.84 & 65.24 & 93.83 \\
$\cP$ & ICNN & ResNet & 87.25 & 82.53 & 85.19 & 83.66 & 65.09 & 93.83 \\
$\cB$ & ICNN & Linear & 88.76 & 81.71 & 87.41 & 82.88 & 53.94 & \textbf{95.06} \\
$\cB$ & ICNN & MLP & 90.20 & 77.98 & 90.37 & 84.97 & 46.83 & 94.44 \\
$\cB$ & ICNN & ResNet & 89.81 & 85.01 & 91.11 & 85.42 & 54.48 & 93.83 \\
\bottomrule
\end{tabular}\label{tab:polytope}
\end{table}

\section{Additional materials}\label{app:materials}

\subsection{Loss functions using other forms of supervision}

To workaround the intractable log-partition function arising in MLE,
another alternative is to use loss functions that leverage other forms of supervision than $(\x,\y)$ pairs. We review them for completeness,
though we emphasize that these methods do \textit{not} learn
the probabilistic model \eqref{eq:ebm}.

\paragraph{Learning from pairwise preferences.}

If we have a dataset of triplets of the form $(\x, \y_+, \y_-)$ 
such that $\y_+ \succ \y_-$ given $\x$, we can learn a Bradley-Terry model
\begin{align}
p_g(\y_1 \succ \y_2|\x) 
&\coloneqq \sigmoid(g(\x, \y_1) - g(\x, \y_2)) \\
&=  \frac{\exp(g(\x, \y_1))}{\exp(g(\x, \y_1)) + \exp(g(\x, \y_2))},
\end{align}
where $\sigmoid(u) \coloneqq 1 / (1 + \exp(u))$ is the logistic sigmoid. The loss associated with the triplet $(\x,\y_+,\y_-)$ is the pairwise logistic loss,
$-\log p_g(\y_+ \succ \y_-|\x)$.
A major advantage of this approach is that there is no log-partition involved.
The DPO loss \citep{rafailov2024direct} can be thought as a generalization of this pairwise logistic loss to non-uniform prior distributions $q$. However, it learns a probabilistic model of 
$\y_1 \succ \y_2$ given $\x$, not of $\y$ given $\x$.

\paragraph{Learning from pointwise scores.}

As an alternative, if we have a dataset of triplets $(\x, \y, t)$ where $t$ is a scalar score assessing the affinity between $\x$ and $\y$, for instance a binary score or a real value score, we can use $\ell(g(\x, \y), t)$ for some classification or regression loss $\ell$.
Again, no log-partition is involved as we can estimate the parameters of $g$ directly. However, this corresponds to learning a probabilistic model of $t$ given $(\x,\y)$, not of $\y$ given $\x$.

\subsection{Fenchel-Young losses}

Let us define some dual regularization
\begin{equation}
\Omega_1(p) \coloneqq \Omega_+(p) + \iota_{\cP_1(\cY)}(p),
\end{equation}
where $\iota_\cS$ is the indicator function of the set $\cS$
and $\dom(\Omega_+) \subseteq \cP_+(\cY)$.
A common choice for $\Omega_+$ is the negative Shannon entropy
$\Omega_+(p) = \langle p, \log p \rangle$
but we will see below how to define very general family of regularizations using $f$-divergences.
Given $g \in \cF(\cY)$ and $p \in \cP_1(\cY)$,
we define the Fenchel-Young loss \citep{blondel2020learning} regularized by $\Omega_1$ as
\begin{align}
L_{\Omega_1}(g, p) 
&\coloneqq  \Omega_1^*(g) + \Omega_1(p)  - \langle g, p \rangle \\
&= \Omega_1^*(g) + \Omega_1(p)  - \EE_{y \sim p}[g(y)].
\label{eq:fy_loss}
\end{align}
The Fenchel-Young loss is primal-dual,
in the sense that we see $g \in \cF(\cY)$ as a primal variable (a function) and $p \in \cP_1(\cY)$ 
as a dual variable (a distribution).
We will omit the constant term $\Omega(p)$ in our derivations.
While we focus on discrete spaces in this paper,
Fenchel-Young losses were also extended to continuous spaces in \citet{mensch2019geometric,martins2022sparse}.

\section{Proofs}\label{app:proof}

We define the shorthand notations 
$\rho_\x \coloneqq \rho(\cdot|\x)$,
$q_\x \coloneqq q(\cdot|\x)$
and
$g_\x \coloneqq g(\x, \cdot)$.

\subsection{Lemmas}

In this section, we state and prove useful lemmas.

\begin{lemma}{$\Omega_1^*$ as unconstrained minimization of $\Omega_+^*$}\label{lemma:dual_simplex}

Let $\Omega_1 \coloneqq \Omega_+ + \iota_{\cP_1(\cY)}$,
where $\Omega_+$ is convex 
with $\dom(\Omega_+) \subseteq \cP_+(\cY)$. 
Then, for all $h \in \cF(\cY)$,
\begin{equation}
\Omega^*_1(h) =
\min_{\tau \in \RR}
\tau + \Omega_+^*(h - \tau) \in \RR
\end{equation}
and
\begin{equation}
\nabla \Omega^*_1(h) =
\nabla \Omega_+^*(h - \tau^\star) \in \cF(\cY),
\end{equation}
where $\tau^\star$ is an optimal solution.
\end{lemma}
\textbf{Proof.}
\begin{align}
\Omega^*_1(h) 
&= \max_{p \in \cP_1(\cY)} \langle h, p \rangle - \Omega_+(p) \\
&= \max_{p \in \cP_+(\cY)} \min_{\tau \in \RR}
\langle h, p \rangle - \Omega_+(p) 
- \tau (\langle p, \ones \rangle - 1) \\
&= \min_{\tau \in \RR} \tau + 
\max_{p \in \cP_+(\cY)} 
\langle h - \tau, p \rangle - \Omega_+(p) \\
&= \min_{\tau \in \RR}
\tau + \Omega_+^*(h - \tau).
\end{align}
The expression of $\nabla \Omega_1^*(h)$ follows from Danskin's theorem. $\square$

\begin{lemma}\label{lemma:conjugate_f_div}
Given $f_+ \colon \RR_+ \to \RR$, $p \in \cP_+(\cY)$ and a fixed reference measure $q \in \cP_+(\cY)$, let
\begin{equation}
\Omega_+(p; q) 
\coloneqq \langle f_+(p/q), q \rangle
= \sum_{\y \in \cY} q(\y) f_+(p(\y)/q(\y)).
\end{equation}
Then,
\begin{equation}
\Omega_+^*(h) = \sum_{\y \in \cY} q(\y) f^*_+(h(\y)) \in \RR
\end{equation}
and
\begin{equation}
\nabla \Omega_+^*(h)(\y) = q(\y) (f_+^*)'(h(\y)) \in \RR_+.
\end{equation}
\end{lemma}
\textbf{Proof.} This follows from classical conjugate calculus.
$\square$

\subsection{Proof of Proposition \ref{prop:min_min_mle} (equivalence with MLE, expected risk setting)}\label{proof:min_min_mle}

Proposition \ref{prop:min_min_mle} is a corollary of Proposition \ref{prop:min_min_fy} with the choice
\begin{equation}
\label{eq:f_kl}
f(u) \coloneqq u \log u - (u - 1) = f_+(u).
\end{equation}
In this case, we obtain
\begin{equation}
f^*(v) = f_+^*(v) = \exp(v) - 1.
\end{equation}
Using Lemma \ref{lemma:conjugate_f_div}, we obtain for all $\x \in \cX$
\begin{equation}
\Omega_+^*(g(\x, \cdot) - \tau(\x); q_\x) = \sum_{\y \in \cY} q(\y|\x) [\exp(g(\x, \y) - \tau(\x)) - 1].
\end{equation}
For $\x$ fixed, minimizing \wrt $\tau$, we get
\begin{align}
\sum_{\y \in \cY} q(\y|\x) (f_+^*)'(g(\x, \y) - \tau(\x)) = 1
& \iff
\sum_{\y \in \cY} q(\y|\x) \exp(g(\x, \y) - \tau(x)) = 1 \\
& \iff 
\sum_{\y \in \cY} q(\y|\x) \exp(g(\x, \y)) = \exp(\tau(\x)) \\
& \iff
\log \sum_{\y \in \cY} q(\y|\x) \exp(g(\x, \y)) = \tau(\x),
\end{align}
which is continuous in $\x$ if $g$ is itself continuous.
Thus the overall objective is
\begin{align}
\min_{g \in \cF(\cX \times \cY)}
\min_{\tau \in \cF(\cX)}
\EE_\x \left[
\tau(\x) + 
\sum_{\y \in \cY} q(\y|\x) [\exp(g(\x, \y) - \tau(\x)) - 1] \right]
-
\EE_{(\x,\y)} g(\x, \y)
\end{align}
and
\begin{equation}
\pi_{g^\star,q}(\y|\x) = q(\y|\x) \exp(g^\star(\x, \y) - \tau(\x))
\end{equation}

\subsection{Proof of Proposition \ref{prop:finite} (optimality in the empirical risk setting)}\label{proof:finite}

Using Lemma \ref{lemma:dual_simplex} with $h \coloneqq g(\x, \cdot)$, we obtain
\begin{align}
\min_{g \in \cF(\cX \times \cY)}
\frac{1}{n}
\sum_{i=1}^n
\left[
\Omega^*_1(g_{\x_i}; q_{\x_i}) - \langle g_{\x_i}, \delta_{\y_i} \rangle\right]
&= \min_{g \in \cF(\cX \times \cY)} 
\frac{1}{n}
\sum_{i=1}^n
\left[
\left(\min_{\tau_i \in \RR} \tau_i + \Omega_+^*(g_{\x_i} - \tau_i; q_{\x_i})\right) - \langle g_{\x_i}, \delta_{\y_i} \rangle\right]. \\
&= \min_{g \in \cF(\cX \times \cY)} 
\min_{\tauv \in \RR^n}
\frac{1}{n}
\sum_{i=1}^n
\left[
\left(\tau_i + \Omega_+^*(g_{\x_i} - \tau_i; q_{\x_i})\right) - \langle g_{\x_i}, \delta_{\y_i} \rangle\right],
\end{align}
where $\delta_\y \in \cP_1(\cY)$ is a delta distribution such that $\delta_\y(\y') = 1$ for $\y' = \y$ and $0$ for $\y' \neq \y$.
Choosing \eqref{eq:f_kl} as in the proof of 
Proposition \ref{prop:min_min_mle}
gives the final result.

\subsection{Proof of Proposition \ref{prop:min_max_mle} (min-max formulation)}
\label{proof:min_max_mle}

We have 
\begin{align}
\min_{g \in \cF(\cX \times \cY)} \cL(g) 
&= \min_{g \in \cF(\cX \times \cY)}
\EE_\x \left[\LSE_g(x) - \langle g_\x, \rho_\x \rangle \right] + \mathrm{const} \\
&= \min_{g \in \cF(\cX \times \cY)}
\EE_\x \left[\left(\max_{p_\x \in \cP_1(\cY)} \langle g_\x, p_\x \rangle - \mathrm{KL}(p_\x, q_\x)\right) - \langle g_\x, \rho_\x \rangle \right] + \mathrm{const}\\
&= \min_{g \in \cF(\cX \times \cY)}
\max_{p \in \cP_1(\cY|\cX)} \EE_\x \left[(\langle g_\x, p_\x \rangle - \mathrm{KL}(p_\x, q_\x)) - \langle g_\x, \rho_\x \rangle \right] + \mathrm{const}\\
&= \min_{g \in \cF(\cX \times \cY)}
\max_{p \in \cP_1(\cY|\cX)} \EE_\x 
\left[
\left(\EE_{\y \sim p_\x}[g(\x, \y)] - \mathrm{KL}(p_\x, q_\x)\right)
-
\EE_{\y \sim \rho_\x}[g(\x, \y)]
\right] + \mathrm{const}.
\end{align}
\subsection{Proof of Proposition \ref{prop:min_min_fy} (optimality in the general Fenchel-Young loss and expected risk setting)}
\label{proof:min_min_fy}

Using Lemma \ref{lemma:dual_simplex} with $h \coloneqq g(\x, \cdot)$, we obtain
\begin{align}
\min_{g \in \cF(\cX \times \cY)}
\EE_\x\left[
\Omega^*_1(g_\x; q_\x) - \langle g_\x,\rho_\x \rangle\right]
&= \min_{g \in \cF(\cX \times \cY)} 
\EE_\x \left[
\left(\min_{\tau_\x \in \RR} \tau_\x + \Omega_+^*(g_\x - \tau_\x; q_\x)\right) - \langle g_\x, \rho_\x \rangle\right].
\end{align}
We are going to show that 
$$
\min_{\tau_\x \in \RR} \tau_\x + \Omega_+^*(g_\x - \tau_\x; q_\x)= \tau^\star(\x) +  \Omega_+^*(g_\x - \tau^\star(\x); q_\x)
$$ where $\tau^\star \in \cF(\cX)$.

\paragraph{Step 1.} We first show that we can define a function $\x \mapsto \tau^\star(\x)$.

We have 
$\tau_\x + \Omega_+^*(g_\x - \tau_\x; q_\x) = \tau_\x + \sum_{y\in \cY} q(\y |\x) f^* (\max\{g(\x, \y) - \tau_\x, f'(0)\})$ with $f'(0) \coloneqq \lim_{x \rightarrow 0, x \geq 0} f'(x) \in \RR \cup \{-\infty\}$. Without loss of generality, we can assume that $q(\y|\x) > 0$ for all $\y$ and $\x$.

Since $(0, +\infty) \subseteq \dom f'$, $q >0$ and $\cY$ is finite, $f'\left(\left(\sum_{y \in \cY} q(\y|\x)\right)^{-1}\right)$ and $f'(\frac1{\min_{\y \in \cY}q({\y|\x})})$ 
are well defined. We can then define
\begin{align*}
    \tau_{\min}(\x) & \coloneqq \max_{\y \in \cY}g(\x, \y) -  f'\left(\frac1{\min_{\y \in \cY}q({\y|\x})}\right) \\
    \tau_{\max}(\x) & \coloneqq \max_{\y \in \cY}g(\x, \y) -  f'\left(\left(\sum_{y \in \cY} q(\y|\x)\right)^{-1}\right). 
\end{align*}
Since $q>0$ and since $f'$ is increasing, one has $\tau_{\min}(\x) < \tau_{\max}(\x)$.

We can then analyze the following function on $[\tau_{\min}(\x), \tau_{\max}(\x)] \times \cX$:
\begin{align*}
    h(\tau, \x) & \coloneqq \tau + \sum_{y \in \cY} q(\y|\x) f^*(\max\{g(\x,\y)-\tau, f'(0)\}).
\end{align*}

First, we need to ensure that we can compute derivatives of this function with respect to its first variable $\tau$ in $[\tau_{\min}(\x), \tau_{\max}(\x)]$.
We have that $ f^*(\max\{g(\x, \y)-\tau, f'(0)\}) = (f+\iota_{\RR_+})^*(g(\x,\y) -\tau)$ and $\dom (f+\iota_{\RR_+})_*' = \dom f_*' \cup (-\infty, f'(0)]$. 
Therefore, if $f'(0) > -\infty$, the domain of $(f+\iota_{\RR_+})_*'$ is unbounded below.
Otherwise, if $f'(0) = -\infty$, since $\mathrm{Im} f' \subseteq \dom f_{*}'$,
the domain of $f_*'$, and so of $(f+\iota_{\RR_+})_*'$, are unbounded below.
Denoting then $\alpha = \sup \dom (f+\iota_{\RR_+})_*'$, 
since $\max_{\y\in \cY}g(\x, \y) - \tau_{\min}(\x) = f'(\frac1{\min_{\y \in \cY}q({\y|\x})}) \in \dom f_*'$, 
we have  $\max_{\y\in \cY}g(\x, \y) - \tau_{\min}(\x) < \alpha$ and therefore
\begin{align*}
    & \tau \geq \tau_{\min}(\x) \\
    \implies & \max_{\y\in \cY}g(\x, \y) - \tau < \alpha \\
    \iff & g(\x, \y) - \tau < \alpha, \mbox{for all} \ \y \in \cY \\
    \implies & \tau \in \dom h'.
\end{align*}
We can then show that $h'(\tau_{\min}(\x), \x) \leq 0$ and $h'(\tau_{\max}(\x), \x) \geq 0$. Indeed, denoting $\y^\star \in \argmax g(\x, \y)$ (we omit the dependency in $x$ to ease the notations), we have
\begin{align*}
    h'(\tau_{\min}(\x), \x) 
    & = 1 - \sum_{\y\in\cY} q(\y|\x) f_*'(\max\{g(\x,\y) - \tau_{\min}(\x), f'(0)\}) \\
    & \stackrel{(i)}{\leq} 
    1 -  q(\y^\star|\x) f_*'(\max\{g(\x,\y^\star) -\tau_{\min}(\x), f'(0)\}) \\
    & = 1 - q(\y^\star|\x)f_*'(\max\{f'(\frac1{\min_{\y \in \cY}q(\y|\x)}), f'(0)\}) \\
    & \stackrel{(ii)}{=} 1 - \frac{q(\y^\star|\x)}{\min_{y\in \cY} (q(\y,\x))} \leq 0,
\end{align*}
where in $(i)$ we used that $q>0$ and $f_*'(\max\{z, f'(0)\}) \geq 0$ for any $z \in \dom (f + \iota_{\RR_+})_*'$, and in $(ii)$, we used that  $f'$ is increasing and $f_*'(f'(p)) = p$ for any $p \in \dom f'$.

Similarly, we have
\begin{align*}
    h'(\tau_{\max}(\x), \x) 
    & = 1 - \textstyle{\sum_{y\in \cY}} q(\y|\x) f_*'(\max\{g(\x,\y) - \tau_{\max}(\x), f'(0)\}) \\
    & \stackrel{(i)}{\geq} 
    1 - \left(\textstyle{\sum_{\y\in\cY}} q(\y|\x)\right) f_*'(\max\{g(\x, \y^\star) -\tau_{\max}, f'(0)\}) \\
    & = 1 -\left(\textstyle{\sum_{\y \in \cY}} q(\y|\x) \right)
    f_*'\left(\max\left\{f'\left(\left(
    \textstyle{\sum_{\y \in \cY}} q(\y|\x)\right)^{-1}\right), 
    f'(0)\right\}\right) \\
    & \stackrel{(ii)}{=} 0,
\end{align*}
where in $(i)$ we used that $\sum_\y a_\y b_\y \leq (\sum_\y a_\y) \max_\y b_\y$ 
if $a_\y \geq 0$ with here $a_\y = q(\y|\x) > 0$ and $b_\y = f_*'(\max\{g(\x, \y) - \tau_{\max}(\x), f'(0)\})$, 
and in $(ii)$ we used the same reasoning as for $h'(\tau_{\min})$.

Finally, we show that $h'$ is increasing on $[\tau_{\min}(\x), \tau_{\max}(\x)]$. When $\tau \leq \tau_{\max}(\x)$, one has $\max_{\y \in \cY}g(\x, \y) - \tau \geq \max_{\y \in \cY}g(\x, \y) - \tau_{\max}(x) \geq f'(0).$

Since $\theta \mapsto f_*'(\max\{\theta, f'(0)\})$ is increasing on $\dom f^* \setminus (-\infty, f'(0)]$ we have that $\tau \mapsto -f_*'(\max\{(g(\x,\y^\star)-\tau, f'(0)\})$ is increasing on $[ \tau_{\min} (\x), \tau_{\max}(\x)]$ and therefore $\tau \mapsto h'(\tau, \x) = 1 - \sum_{\y \in \cY} q(\y|\x) f_*'(\max\{g(\x, \y) - \tau, f'(0)\})$ is increasing on $[\tau_{\min}(\x), \tau_{\max}(\x)]$.

Overall $h$ is well defined and strictly convex on $[\tau_{\min}(\x), \tau_{\max}(\x)]$ such that $h'(\tau_{\min}(\x)) \leq 0$ and $h'(\tau_{\max}(\x)) \geq 0$. Hence, since $h$ is convex, we have 
\[
\inf_{\tau \in \RR} h(\tau, \x) = \min_{\tau_{\min}(\x) \leq \tau \leq \tau_{\max}(\x)} h(\tau, \x)
\]
and the unique minimizer $\tau^\star(\x)$ can then be found by solving the first order optimality condition $h'(\tau, \x) = 0$ in $[\tau_{\min}(\x), \tau_{\max}(\x)]$. We can therefore define a function $\tau^\star$ mapping each $\x$ to the unique root of $h'(\tau,\x)$ in $[\tau_{\min}(\x), \tau_{\max}(\x)]$. 

\paragraph{Step 2.} We then show that $\tau^\star$ is continuous.

Note that, because $\cY$ is finite and $\x \mapsto q(\y|\x)$, $\x \mapsto g(\x,\y)$ and $f'$ are continuous, both $\tau_{\min}$ and  $\tau_{\max}$ are continuous functions of $\x$. Let $(\x_k)_k$ be a sequence such that $\x_k \to \x$ as $k \to +\infty$. Since  $\tau_{\min}$ and  $\tau_{\max}$ are continuous, $\tau_{\min}(\x_k) \to \tau_{\min}(\x)$ and $\tau_{\max}(\x_k) \to \tau_{\max}(\x)$. Therefore, $(\tau^\star(\x_k))_k$ is a bounded sequence. As such, it admits a converging sub-sequence $(\tau^\star(\x_{\varphi(k)}))_k$ converging to some $t^\star(\x)$. One has
$
h'(\tau^\star(\x_{\varphi(k)}),\x_{\varphi(k)}) = 0
$. By continuity of $h'$, this gives $h'(t^\star(\x),\x) = 0$
and therefore $ t^\star(\x) = \tau^\star(\x)$ by unicity of the root. Therefore, any converging subsequence of the bounded sequence $(\tau^\star(\x_k))$ converges to $\tau^\star(\x)$. This exactly proves that $\tau^\star(\x_k) \to \tau^\star(\x)$ as $k \to +\infty$. Therefore, $\tau^\star$ is continuous and
\begin{align}
\EE_\x \min_{\tau_\x \in \RR }
\left[
\tau_\x + \Omega_+^*(g_\x - \tau_\x)\right] = \EE_\x \left[\tau^\star(\x) + \Omega_+^*(g_\x - \tau^\star(\x))\right].
\end{align}

\paragraph{Step 3.} Last, we show that $\min_{\tau \in \cF(\cX)}\EE_\x \left[\tau(\x) + \Omega_+^*(g_\x - \tau(\x))\right] = \EE_{\x} \min_{\tau_\x \in \RR} \left[\tau_\x + \Omega_+^*(g_\x - \tau_\x)\right].$

By definition of $\tau^\star$, we have $\min_{\tau \in \cF(\cX)}\EE_\x \left[\tau(\x) + \Omega_+^*(g_\x - \tau(\x))\right] \leq \EE_\x \left[\tau^\star(\x) + \Omega_+^*(g_\x - \tau^\star(\x))\right] = \EE_\x \min_{\tau_\x \in \RR }
\left[
\tau_\x + \Omega_+^*(g_\x - \tau_\x)\right]$.
On the other hand, for any $x$ and any $\tau \in \cF(\cX)$, we have 
$$\EE_\x \left[\tau(\x) + \Omega_+^*(g_\x - \tau(\x))\right] \geq \EE_{\x} \min_{\tau_\x \in \RR} \left[\tau_\x + \Omega_+^*(g_\x - \tau_\x)\right].$$
Since this holds for any continuous function $\tau$, this gives 
$$\min_{\tau \in \cF(\cX)}\EE_\x \left[\tau(\x) + \Omega_+^*(g_\x - \tau(\x))\right] \geq \EE_{\x} \min_{\tau_\x \in \RR} \left[\tau_\x + \Omega_+^*(g_\x - \tau_\x)\right].$$
We then have 
$$\min_{\tau \in \cF(\cX)}\EE_\x \left[\tau(\x) + \Omega_+^*(g_\x - \tau(\x))\right] = \EE_{\x} \min_{\tau_\x \in \RR} \left[\tau_\x + \Omega_+^*(g_\x - \tau_\x)\right].$$
Therefore, 
\begin{align}
\min_{g \in \cF(\cX \times \cY)}
\EE_\x\left[
\Omega^*_1(g_\x; q_\x) - \langle g_\x,\rho_\x \rangle\right]
&= \min_{g \in \cF(\cX \times \cY)}
\min_{\tau \in \cF(\cX)}
\EE_\x \left[
\tau(\x) + \Omega_+^*(g_\x - \tau(\x)) - \langle g_\x,\rho_\x \rangle\right].
\end{align}

In addition, for all $(\x,\y) \in (\cX \times \cY)$
\begin{equation}
\pi_{g^\star,q}(\y|\x) 
= \nabla \Omega_+^*(g^\star_\x - \tau^\star(\x); q_\x)(\y)
=  q(\y|\x) (f_+^*)'(g^\star(\x, \y)). 
\end{equation}
This concludes the proof.
\subsection{Joint convexity}
\label{proof:joint_convexity}

\begin{proposition}
Given $q\in \cP_+(\cY|\cX)$, $f$ differentiable, strictly convex well defined on $(0, +\infty)$, for a coupling $g_{\w}(x, y)$ linear in $\w \in \cW$ and a finite dimensional representation of the log-partition function, $\tau_{\v}(\x_i) = v_i$ for $i \in [n]$, the objective
\begin{align*}
\widetilde{\cL}_{f}(g_{\w}, \tau_{\v}) 
& \coloneqq 
\frac{1}{n} \left(\sum_{i=1}^n L^f_{g_{\w},\tau_{\v}}(\x_i) - g_{\w}(\x_i, \y_i)\right),\\ 
\mbox{where} \quad L_{g_{\w}, \tau_{\v}}^f(\x_i)
& \coloneqq
\tau_{\v}(\x_i) + \sum_{\y' \in \cY} q(\y'|\x_i) f_+^*(g_{\w}(\x_i, \y') - \tau_{\v}(\x_i))
\end{align*}
is jointly convex in $\w, \v$, where $f_+^*$ is the convex conjugate of the restriction of $f$ to $\RR_+$.
\end{proposition}
\begin{proof}
The function $(\w, \v_ \mapsto g_{\w}(\x_i, \y') - \tau_{\v}(\x_i)$ is linear in $\w, \v$ by assumption.
Since $f_+^*$ is convex, $(\w, \v) \mapsto f_+^*(g_{\w}(\x_i, \y') - \tau_{\v}(\x_i))$ is then jointly convex in $\w, \v$ 
since the composition of a linear function and a convex function is always convex.
Since $q(\y'|\x_i)$ is positive, $(\w, \v) \mapsto q(\y'|\x_i)f_+^*(g_{\w}(\x_i, \y') - \tau_{\v}(\x_i))$ is also jointly convex.
The functions $\w \rightarrow g_{\w}(\x_i, \y_i)$ and $\v \mapsto \tau_{\v}(\x_i)$ are linear in $\w$ and $\v$ respectively.
A sum of convex functions is convex hence the result.
\end{proof}

\subsection{Proof of gradient computation}
\label{proof:gradient_estimator}

One has 
\begin{align}
\nabla_\w \LSE_{g_\w}(\x) 
&= \nabla_\w \log \sum_{\y \in \cY} q(\y|\x) \exp(g_\w(\x,\y)) \\
&= \frac{\sum_{\y \in \cY} q(\y|\x) \nabla_\w \exp(g_\w(\x,\y))}{\sum_{\y \in \cY} q(\y|\x) \exp(g_\w(\x,\y))} \\
&= \frac{\sum_{\y \in \cY} q(\y|\x)  \exp(g_\w(\x,\y)) \nabla_\w g_\w(\x, \y)}{\sum_{\y \in \cY} q(\y|\x) \exp(g_\w(\x,\y))} =  \frac{\EE_{\y \sim q(\cdot|\x)}\left[\exp(g_\w(\x,\y)) \nabla_\w g_\w(\x, \y)\right]}{\EE_{\y \sim q(\cdot|\x)}\left[\exp(g_\w(\x,\y))\right]} \\
&= \sum_{\y \in \cY} p_{g_\w}(\y|\x) \nabla_\w g_\w(\x, \y) = \EE_{\y \sim p_{g_\w}(\cdot|\x)}\left[\nabla_\w g_\w(\x, \y)\right].
\end{align}

\end{document}